\documentclass[journal]{IEEEtran}

\usepackage{amsmath,amsfonts,amssymb}
\usepackage{algorithm}
\usepackage{algpseudocode}
\usepackage{array}
\usepackage{bbding}
\usepackage{bm}
\usepackage{booktabs}
\usepackage{cite}
\usepackage{graphicx}
\usepackage{mathrsfs}
\usepackage{multirow}
\usepackage[caption=false,font=footnotesize]{subfig}
\usepackage{textcomp}
\usepackage{threeparttable}
\usepackage{url}
\usepackage[table]{xcolor}
\usepackage{hyperref}
\graphicspath{{../}}
\renewcommand{\eqref}[1]{Eq.~(\ref{#1})}

\definecolor{tblBg}{HTML}{FFF1CB}
\definecolor{tblGroup}{HTML}{FAE4D5}
\definecolor{tblHead}{HTML}{FFF6D8}
\definecolor{tblLine}{HTML}{2B2B2B}
\definecolor{mygreen}{RGB}{112,173,71}
\definecolor{mygray}{gray}{.9}

\title{UniPolar: Joint Image and Polarimetric Restoration under Compound Degradations via Physical Consistency Learning}
\author{Yang Jin, Guangyu Guo, Linghao Shen, Binglu Wang$\dagger$
\thanks{Yang Jin and Binglu Wang are with School of Astronautics, Northwestern Polytechnical University, Xi'an 710072, China (e-mail: \{jin91999, wbl921129\}@gmail.com)}
\thanks{Guangyu Guo is with School of Computer Science and Technology, Zhejiang University, Hangzhou 310058, China (e-mail: gyguo95@gmail.com)}
\thanks{Linghao Shen is with School of Marine Science and Technology, Tianjin University, Tianjin, 300072, China (e-mail: shenlinghao@tju.edu.cn)}
\thanks{$\dagger$Corresponding author:Binglu Wang}
}

\begin{document}
\maketitle
\begin{abstract}
Polarimetric image restoration aims to recover polarization-angle images and reliable polarimetric quantities from degraded observations. Beyond visual fidelity, it also requires preserving the physical relationships underlying reliable polarimetric estimation. However, existing methods typically address either device- or environment-induced degradations in isolation, limiting their applicability to color-polarization filter array (CPFA) imaging, where both often coexist. Such compound degradations jointly perturb image intensities and cross-angle polarization differences. Consequently, improved image fidelity alone does not necessarily ensure accurate Stokes, DoLP, and AoLP recovery. To address this challenge, we propose UniPolar, a coupled Image–Stokes framework for joint image–polarimetric restoration from compound-degraded CPFA observations. Specifically, UniPolar employs two complementary streams to reconstruct four polarization-angle images and directly predict Stokes difference components. To coordinate these streams, Degradation-Conditioned Cross-Domain Interaction combines Dual-Domain Frequency Interaction for frequency-selective evidence exchange with Degradation-Conditioned Modulation for condition-adaptive feature correction. Furthermore, at the output level, Image–Stokes Physical Consistency Learning aligns image-derived and directly predicted Stokes components while regularizing angular fidelity and the physical feasibility of DoLP. On the compound-degradation benchmark, UniPolar achieves average image and DoLP PSNRs of 31.522 dB and 24.097 dB, respectively, with an average AoLP error of 16.833°. These results underscore the importance of jointly considering visual fidelity and polarimetric reliability, offering a physically grounded perspective on learning-based restoration under compound degradations.
\end{abstract}


\section{Introduction}
\label{sec:intro}
Polarimetric imaging captures physical cues beyond RGB intensity and has been widely used in reflection removal, reflective-surface reconstruction, and shape recovery~\cite{yao2025polarfree,wu2025glossy,luo2025cpifuse,Shao2022PolarimetricIR,Wu2021SingleImageSH}. In color-polarization filter array (CPFA) imaging systems, polarized images with different polarizer angles are spatially multiplexed on a single sensor, from which Stokes parameters, the degree of linear polarization (DoLP), and the angle of linear polarization (AoLP) are estimated. Therefore, polarimetric image restoration is not merely a multi-channel image reconstruction problem: it must recover visually faithful polarized images while preserving the physical relationships required for reliable AoLP and DoLP estimation.

\begin{figure}[!t]
\centering
\includegraphics[width=1\linewidth]{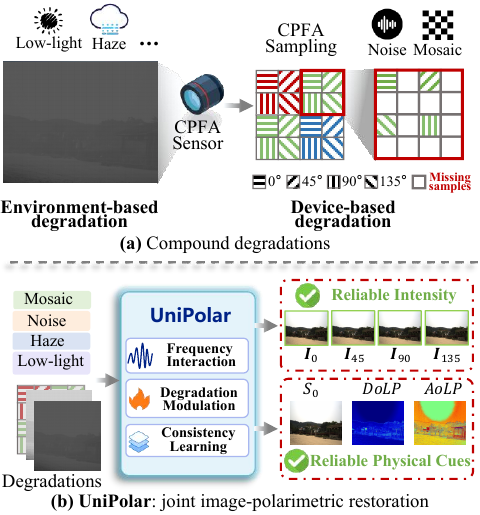}
\caption{(a) Coupled device- and environment-induced degradations in CPFA observations impair both image and polarimetric restoration. (b) UniPolar jointly recovers polarization-angle images and polarimetric quantities, promoting Image--Stokes consistency from cross-domain representations to physically constrained outputs.}

\label{fig:intro}
\end{figure}

In CPFA polarimetric imaging, observations are usually affected by compound device-based and environment-based degradations, as illustrated in Fig.~\ref{fig:intro}(a). On the device side, the spatial multiplexing of different polarizer angles leads to directional undersampling, producing a mosaicked observation, while sensor acquisition introduces resolution loss and noise. On the environment side, adverse imaging conditions such as low illumination and haze reduce contrast, degrade polarization measurements, and lower the effective signal-to-noise ratio. These degradations often occur jointly, producing sparse and noisy CPFA observations under challenging illumination or scattering conditions. They not only reduce visual quality but also disrupt the Stokes-consistent relationships among polarizer-angle measurements, with errors propagating to Stokes parameters and derived polarization attributes.

Existing task-specific polarimetric image restoration methods have advanced individual tasks, including demosaicking, denoising, low-light enhancement, deblurring, super-resolution, and dehazing~\cite{zhang2018learning,wen2019convolutional,qiu2021linear,nguyen2022two,liu2023polarization,zhou2023polarization,dong2022polarization,zhou2024learning,rahman2025polarization,li2026polarization,zhou2025pidsr,li2025demosaicking}. Most, however, primarily address device-induced degradations or isolated environmental conditions, without sufficiently accounting for their compounded effects. Recent joint and shared-architecture methods broaden task coverage through joint demosaicking, denoising, super-resolution, or Image--Stokes modeling~\cite{zhou2025pidsr,rahman2025polarization,Zhang2024PolarizationBasedTI,Zhou2026ArchitecturalUF}, but they are generally optimized for predefined degradation pipelines or require degradation-specific training. General multi-degradation restoration methods instead handle multiple degradation types within one model~\cite{Li2022AllInOneIR,Valanarasu2022TransWeather,Park2023AllinOneIR,Potlapalli2023PromptIRPF,Ma2023ProRes,Luo2024ControllingVL,Conde2024InstructIR,Guo2024OneRestore,Cui2025AdaIR,Wu2025TUR,Tian2025DFPIR,Luo2025VisualInstructed,Jiang2024ASO}. Nevertheless, these methods mainly target conventional RGB images without explicitly modeling cross-angle polarization differences or the physical relationships required for reliable Stokes estimation. These limitations highlight the challenge of jointly restoring visual fidelity and reliable polarimetric information from compound-degraded CPFA observations.

To bridge this gap, we formulate compound-degraded CPFA restoration as a physically coupled image–polarimetric restoration problem that requires simultaneously recovering visually faithful polarization-angle images and reliable polarimetric quantities from the same degraded observation. To this end, we propose UniPolar, a coupled Image--Stokes framework illustrated in Fig.~\ref{fig:intro}(b). UniPolar preserves image and polarimetric representations in separate streams and couples them through Degradation-Conditioned Cross-Domain Interaction, where Dual-Domain Frequency Interaction (DDFI) exchanges frequency-selective evidence and Degradation-Conditioned Modulation (DCM) applies condition-adaptive corrections. At the output level, Image--Stokes Physical Consistency Learning (PCL) reconciles image-derived and directly predicted polarimetric quantities through component, angular, and feasibility constraints. This design provides a physically informed approach to compound-degraded CPFA restoration, emphasizing that polarimetric reliability should be explicitly considered alongside image fidelity.

Our main contributions are summarized as follows:
\begin{itemize}

\item We formulate compound-degraded CPFA restoration as a physically coupled problem, aiming to jointly recover visually faithful polarization-angle images and reliable polarimetric quantities.

\item We propose UniPolar, a coupled Image--Stokes framework with Degradation-Conditioned Cross-Domain Interaction, combining DDFI and DCM for frequency-selective evidence exchange and condition-adaptive feature correction.

\item We introduce Image–Stokes Physical Consistency Learning to align image-derived and directly predicted polarimetric quantities through component, angular, and feasibility constraints, thereby improving physical consistency and restoration accuracy.
\end{itemize}

\section{Related Work}
\label{sec:related_work}

\subsection{Polarimetric Image Restoration}
Snapshot division-of-focal-plane (DoFP) and CPFA cameras spatially multiplex analyzer-angle measurements, requiring full-resolution polarized images to be reconstructed from sparse samples. Classical methods use edge-aware or physically motivated interpolation~\cite{ratliff2009interpolation,Morimatsu2020MonochromeAC,Morimatsu2021MonochromeAC} and sparse priors~\cite{zhang2018sparse}. Learning-based approaches progress from convolutional and linear-polarization formulations~\cite{wen2019convolutional,qiu2021linear} to two-stage, jointly regularized, and diffusion-based reconstruction~\cite{nguyen2022two,liu2023polarization,luo2024learning,li2025demosaicking,li2026polarization}.

Beyond demosaicking, polarization-aware priors support supervised and self-supervised denoising~\cite{li2020learningdenoising,li2023polarized,liu2023pol2pol}, super-resolution and burst reconstruction~\cite{yu2023color,hwang2025benchmarking}, and restoration under low light, motion blur, or scattering~\cite{zhou2023polarization,zhou2024learning,Shen2019AnII,Zhou2021LearningTD,dong2022polarization,Zhang2024PolarizationBasedTI}. Recent systems further combine demosaicking with super-resolution or denoising, or share architectures across multiple polarimetric degradations~\cite{zhou2025pidsr,rahman2025polarization,Zhou2026ArchitecturalUF}. Most, however, assume one degradation family, a fixed pipeline, or degradation-specific training, without jointly modeling compound device- and environment-induced corruption and reconciling image and Stokes outputs. UniPolar addresses this gap through a single jointly trained model with Image--Stokes consistency at both representation and output levels.

\begin{figure*}[!t]
\centering
\includegraphics[width=1\linewidth]{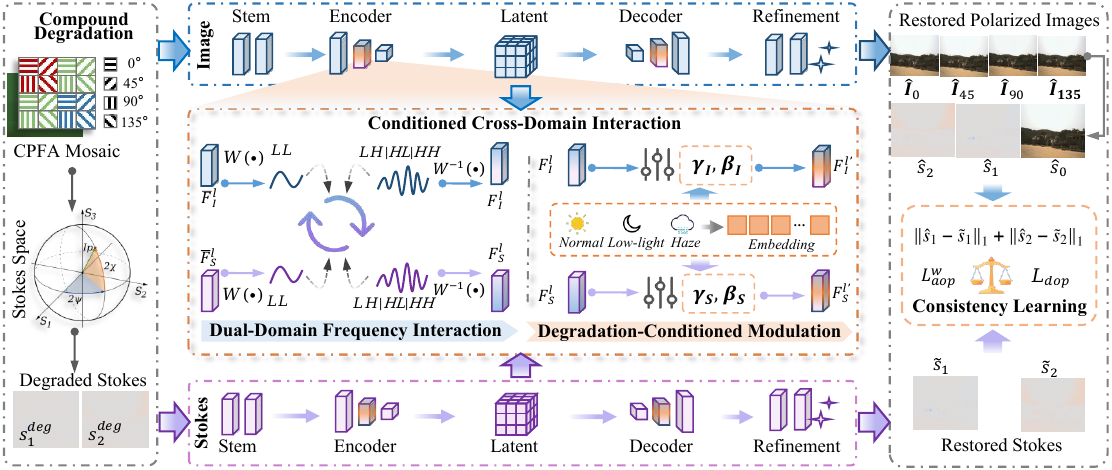}
\caption{\textbf{Overview of UniPolar.} Given compound-degraded CPFA samples and their masks, the image stream reconstructs four polarization-angle images, while the Stokes stream estimates the difference components $s_1$ and $s_2$ from degraded polarimetric cues. DDFI exchanges complementary low- and high-frequency information between the streams, and DCM adapts the coupled features to the degradation condition. Image--Stokes PCL finally reconciles image-derived and directly predicted Stokes quantities through component, angular, and DoLP-feasibility constraints, yielding visually faithful images and physically reliable polarimetric outputs.}
\label{fig:framework}
\end{figure*}

\subsection{Multi-Degradation Image Restoration}
Multi-degradation image restoration aims to handle multiple degradation types and their combinations within a unified model~\cite{Jiang2024ASO}. Early methods infer degradation representations from corrupted inputs or employ adaptive filters to specialize shared restoration features~\cite{Li2022AllInOneIR,Park2023AllinOneIR}. PromptIR learns degradation-aware latent prompts for blind restoration~\cite{Potlapalli2023PromptIRPF}, while scene descriptors or degradation-type prompts provide condition-specific guidance~\cite{Guo2024OneRestore,Tian2025DFPIR}. Recent studies further explore adaptive self-prompting for mixed weather degradations and contrastive learning transferable across dehazing, deraining, and desnowing~\cite{Wen2025AdaptiveSelfPrompting,Cheng2024ProgressiveNegative}. Frequency-aware architectures improve transfer across restoration tasks through vector decomposition or prioritized sparse attention~\cite{Jiang2026VDMamba,Dong2026HFPSAN}, while dynamic reparameterization and degradation-aware state-space prompts reduce task interference within a single model~\cite{Li2026DRNet,Liu2026DPMambaIR}. Related methods also consider composite degradations and frequency-aware feature adaptation~\cite{Guo2024OneRestore,Cui2025AdaIR,Ren2026AnyIR}. Despite their effectiveness on conventional RGB images, these methods remain image-centric and do not explicitly preserve the physical relationships among polarized images at different polarizer angles and their derived Stokes components. UniPolar also conditions a single jointly trained model on the degradation setting, but differs by promoting Image--Stokes compatibility at the representation level and explicitly enforcing their physical consistency at the output level.

\section{Methodology}
\label{sec:method}
\subsection{Problem Formulation and Design Principle}
We formulate compound-degraded CPFA restoration as a joint image and polarimetric restoration problem. From the same degraded observation, image restoration recovers the polarization-angle image set $\mathbf{I}=\{I_0,I_{45},I_{90},I_{135}\}$, while polarimetric restoration recovers the associated Stokes components and reliable DoLP/AoLP. Because these polarimetric quantities are determined by cross-angle image differences, the two objectives are physically coupled rather than independent. Given compound device- and environment-induced degradations, UniPolar takes sparse measurements $\mathbf{X}$, sampling masks $\mathbf{M}$, degraded $s_1/s_2$ cues $\mathbf{s}_{\Delta}^{\mathrm{deg}}$, and an environmental-degradation label $d$ as input:
\begin{equation}
(\hat{\mathbf{I}},\tilde{\mathbf{s}}_{\Delta})
=
\mathcal{F}_{\Theta}(\mathbf{X},\mathbf{M},\mathbf{s}_{\Delta}^{\mathrm{deg}},d).
\end{equation}
Here, $\Theta$ denotes the model parameters, $\hat{\mathbf{I}}=\{\hat{I}_{0},\hat{I}_{45},\hat{I}_{90},\hat{I}_{135}\}$ is the restored image set, and $\tilde{\mathbf{s}}_{\Delta}=\{\tilde{s}_{1},\tilde{s}_{2}\}$ contains the directly predicted Stokes difference components.

For a polarized image set $\mathbf{I}=\{I_0,I_{45},I_{90},I_{135}\}$,
the first three Stokes components $s_0$, $s_1$, and $s_2$ for linear polarization, together with DoLP $p$ and AoLP $\theta$, are computed as:
\begin{equation}
\label{eq:polarimetric_relations}
\begin{gathered}
s_0=I_0+I_{90},\quad
s_1=I_0-I_{90},\quad
s_2=I_{45}-I_{135},\\
p=\frac{\sqrt{s_1^2+s_2^2}}{s_0+\epsilon},\qquad
\theta=\frac{1}{2}\operatorname{atan2}(s_2,s_1),
\end{gathered}
\end{equation}
where $\epsilon$ ensures numerical stability. These relations apply to both clean and restored image sets, with hats denoting quantities derived from $\hat{\mathbf{I}}$. Thus, $\hat{\mathbf{I}}$ and $\tilde{\mathbf{s}}_{\Delta}$ provide image-derived and independently predicted descriptions of the same polarimetric state.

\eqref{eq:polarimetric_relations} provides two estimates of the same Stokes state: one derived from the restored polarization-angle images and one predicted directly. Compound degradations jointly perturb image intensities and cross-angle polarization differences, so independently optimizing image fidelity and polarimetric accuracy need not make these estimates agree. UniPolar therefore uses coupled Image--Stokes representations: separate streams preserve image and polarimetric evidence, degradation-conditioned frequency exchange links them, and output constraints reconcile their components, direction, and feasible magnitude. This design connects joint representation learning directly to the polarimetric formation relations.

\subsection{Coupled Image--Stokes Representation}
The sparse CPFA measurements $\mathbf{X}$ contain the observed RGB samples at four analyzer angles, while the masks $\mathbf{M}$ identify which samples were actually measured. Encoding them together lets the image stream distinguish missing values from valid dark regions and retain the spatial evidence needed to reconstruct each polarization-angle image. The degraded Stokes cues $\mathbf{s}_{\Delta}^{\mathrm{deg}}$ are obtained from mask-aware locally filled observations using the cross-angle differences in \eqref{eq:polarimetric_relations}. They expose a coarse polarization pattern, but also inherit errors from interpolation and the compound degradation. We therefore use these cues to initialize a separate Stokes stream rather than treating them as restored components:
\begin{equation}
\mathbf{F}_{I}=\mathcal{E}_{I}^{\mathrm{stem}}([\mathbf{X},\mathbf{M}]),
\qquad
\mathbf{F}_{S}=\mathcal{E}_{S}^{\mathrm{stem}}(\mathbf{s}_{\Delta}^{\mathrm{deg}}),
\end{equation}
where $[\cdot,\cdot]$ denotes concatenation, and $\mathcal{E}_{I}^{\mathrm{stem}}$ and $\mathcal{E}_{S}^{\mathrm{stem}}$ are the image and Stokes embedding layers, respectively. As shown in Fig.~\ref{fig:framework}, both streams follow encoder--decoder pathways, with features $\mathbf{F}_{I}^{l}$ and $\mathbf{F}_{S}^{l}$ at level $l$. The image stream preserves intensity, boundaries, and texture for restoring $\hat{\mathbf{I}}$; the Stokes stream models the spatial variation of $s_1/s_2$ for predicting $\tilde{\mathbf{s}}_{\Delta}$. Their features describe the same scene at corresponding levels and can therefore exchange complementary evidence, while the separate output pathways retain the distinct image and polarimetric restoration objectives.

\subsection{Degradation-Conditioned Cross-Domain Interaction}
Useful cross-domain evidence depends on both frequency and degradation: global illumination and scattering differ from edges, sampling artifacts, and local polarization variations, while a correction suitable for low light may be inappropriate for haze. Direct, fixed fusion ignores both distinctions. UniPolar therefore uses Dual-Domain Frequency Interaction (DDFI) for frequency-dependent exchange and Degradation-Conditioned Modulation (DCM) for condition-dependent correction.

\noindent\textbf{Dual-Domain Frequency Interaction.}
At each level, DDFI applies bidirectional residual fusion to $\mathbf{F}_{I}^{l}$ and $\mathbf{F}_{S}^{l}$. Stokes-derived channel and spatial gates modulate the image features, while the reverse path injects image structures into the Stokes features. The residual paths retain domain-specific information. The fused features $\bar{\mathbf{F}}_{I}^{l}$ and $\bar{\mathbf{F}}_{S}^{l}$ are then decomposed into wavelet subbands for frequency-specific interaction.

Low illumination and haze are often reflected in low-frequency variations in global intensity and contrast~\cite{xu2020learning,cui2023image,Liu2025FrequencyDD}, whereas CPFA artifacts, noise, edges, and local polarization variations contain substantial high-frequency components~\cite{zhou2025pidsr,zhou2023low}. We therefore decompose each fused feature before frequency-specific interaction, separating global degradation correction from polarization-sensitive detail transfer:
\begin{equation}
\mathcal{W}(\bar{\mathbf{F}}_{m})
=
\{\mathbf{F}_{m}^{LL},\mathbf{F}_{m}^{LH},\mathbf{F}_{m}^{HL},\mathbf{F}_{m}^{HH}\},
\quad m\in\{I,S\}.
\end{equation}
Let $\bar{m}$ denote the domain complementary to $m$, namely $(m,\bar{m})\in\{(I,S),(S,I)\}$. For compactness, we omit the level index $l$ below. The low-frequency bands exchange global information through bidirectional $1\times1$ projections $\mathcal{P}_{\bar{m}\rightarrow m}^{L}$:
\begin{equation}
\mathbf{H}_{m}^{LL}
:=
\mathbf{F}_{m}^{LL}
+\rho_{m}^{L}
\mathcal{P}_{\bar{m}\rightarrow m}^{L}(\mathbf{F}_{\bar{m}}^{LL}),
\quad m\in\{I,S\},
\end{equation}
where $\rho_{m}^{L}$ is a learnable residual scale. For each high-frequency band $b\in\{LH,HL,HH\}$, a difference-aware gate is generated from both domains and their absolute discrepancy:
\begin{equation}
\mathbf{G}_{m}^{b}
:=
\operatorname{sigmoid}\!\left(
\mathcal{G}_{m}^{H}
\left[
\mathbf{F}_{m}^{b},
\mathbf{F}_{\bar{m}}^{b},
\left|\mathbf{F}_{m}^{b}-\mathbf{F}_{\bar{m}}^{b}\right|
\right]
\right).
\end{equation}
The gated cross-domain correction is then applied as:
\begin{equation}
\mathbf{H}_{m}^{b}
:=
\mathbf{F}_{m}^{b}
+\rho_{m}^{H}\,
\mathbf{G}_{m}^{b}\odot
\mathcal{T}_{\bar{m}\rightarrow m}^{H}(\mathbf{F}_{\bar{m}}^{b}),
\end{equation}
where $\rho_m^H$ is a learnable residual scale, $\mathcal{G}_{m}^{H}$ uses two $1\times1$ projections with GELU and sigmoid activations, and $\mathcal{T}_{\bar{m}\rightarrow m}^{H}$ uses depthwise and pointwise convolutions. Both operators are shared across the three directional bands. The interaction-enhanced features are reconstructed by:
\begin{equation}
\mathbf{F}_{m}^{l}
:=
\mathcal{W}^{-1}
\left(
\mathbf{H}_{m}^{LL},
\mathbf{H}_{m}^{LH},
\mathbf{H}_{m}^{HL},
\mathbf{H}_{m}^{HH}
\right),
\end{equation}
where $\mathcal{W}^{-1}(\cdot)$ denotes the inverse wavelet transform and reconstructs the interaction-enhanced feature from the four updated subbands. Restricting the exchange to corresponding bands allows the two streams to share global degradation cues and polarization-sensitive details without collapsing them into an undifferentiated representation.

To carry this frequency-aware modeling into output supervision, we constrain the Laplacian responses of the image-derived total intensity and the predicted Stokes difference components. Let $\hat{s}_0=\hat{I}_0+\hat{I}_{90}$, let $\tilde{s}_k$ denote the Stokes-stream prediction for $k\in\{1,2\}$, and let $s_0,s_1,s_2$ be their clean references. The frequency loss is
\begin{equation}
\begin{aligned}
\mathcal{L}_{\mathrm{freq}}
&=\lambda_{f0}\operatorname{MSE}
\left(\left|\nabla^2\hat{s}_0\right|,
      \left|\nabla^2s_0\right|\right)\\
&\quad+\frac{\lambda_{f12}}{2}
\sum_{k=1}^{2}\operatorname{MSE}
\left(\left|\nabla^2\tilde{s}_k\right|,
      \left|\nabla^2s_k\right|\right),
\end{aligned}
\end{equation}
where $\nabla^2$ denotes a channel-wise $3\times3$ Laplacian operation with one-pixel padding, and $|\cdot|$ is the element-wise absolute value. The index $k$ enumerates the $s_1$ and $s_2$ components, while $\operatorname{MSE}(\cdot,\cdot)$ averages the squared difference over all pixels and color channels. The weights $\lambda_{f0}$ and $\lambda_{f12}$ balance the total-intensity and Stokes-component terms. Matching response magnitudes makes the loss insensitive to the Laplacian sign while preserving intensity boundaries in $s_0$ and polarization-sensitive local variations in $s_1/s_2$.

\noindent\textbf{Degradation-Conditioned Modulation.}
Frequency-selective exchange remains condition agnostic. DCM therefore encodes the degradation label $d$ separately for domain $m\in\{I,S\}$ and projects it into channel-wise scaling and bias parameters. These parameters modulate a compact residual feature, producing a condition-specific correction while retaining the shared representation.

Specifically, at level $l$ of domain $m\in\{I,S\}$, the degradation embedding is defined as $\mathbf{e}_{d,m}^{l}=\operatorname{Emb}_{m}^{l}(d)$. Given an input feature $\mathbf{F}_{m}^{l}$, a bottleneck transformation first extracts the residual representation:
\begin{equation}
\mathbf{Z}_{m}^{l}
:=
\operatorname{GELU}\!\left(
\mathcal{D}_{m}^{l}\!\left(
\operatorname{GELU}\!\left(\mathcal{P}_{m,\mathrm{down}}^{l}(\mathbf{F}_{m}^{l})\right)
\right)
\right),
\end{equation}
where $\mathcal{P}_{m,\mathrm{down}}^{l}$ is a $1\times1$ channel projection and $\mathcal{D}_{m}^{l}$ is a $3\times3$ depthwise convolution. The degradation embedding then generates channel-wise parameters for internal bottleneck modulation:
\begin{equation}
\begin{aligned}
(\boldsymbol{\gamma}_{d,m}^{l,\mathrm{in}},\boldsymbol{\beta}_{d,m}^{l,\mathrm{in}})
&=\mathcal{G}_{m}^{l,\mathrm{in}}(\mathbf{e}_{d,m}^{l}),\\
\mathcal{A}_{m}^{l}(\mathbf{F}_{m}^{l},d)
&=\mathcal{P}_{m,\mathrm{up}}^{l}\!\left[
(1+\boldsymbol{\gamma}_{d,m}^{l,\mathrm{in}})\odot\mathbf{Z}_{m}^{l}
+\boldsymbol{\beta}_{d,m}^{l,\mathrm{in}}
\right],
\end{aligned}
\end{equation}
where $\mathcal{P}_{m,\mathrm{up}}^{l}$ projects the bottleneck feature back to its original channel dimension. A second condition projection produces output-level parameters:
\begin{equation}
(\boldsymbol{\gamma}_{d,m}^{l,\mathrm{out}},\boldsymbol{\beta}_{d,m}^{l,\mathrm{out}})
=\mathcal{G}_{m}^{l,\mathrm{out}}(\mathbf{e}_{d,m}^{l}),
\end{equation}
and the modulated feature is computed as:
\begin{equation}
\begin{aligned}
\mathbf{F}_{m}^{l\prime}
:={}&\mathbf{F}_{m}^{l}
+\alpha_{m,\mathrm{conv}}^{l}\mathcal{A}_{m}^{l}(\mathbf{F}_{m}^{l},d)\\
&+\alpha_{m,\mathrm{FiLM}}^{l}
\left[
\boldsymbol{\gamma}_{d,m}^{l,\mathrm{out}}\odot\mathcal{A}_{m}^{l}(\mathbf{F}_{m}^{l},d)
+\boldsymbol{\beta}_{d,m}^{l,\mathrm{out}}
\right].
\end{aligned}
\end{equation}
where $\mathcal{G}_{m}^{l,\mathrm{in}}$ and $\mathcal{G}_{m}^{l,\mathrm{out}}$ are lightweight conditioning projections, and $\odot$ denotes channel-wise modulation. The independent scales $\alpha_{m,\mathrm{conv}}^{l}$ and $\alpha_{m,\mathrm{FiLM}}^{l}$ control the convolutional residual and output-level modulation, respectively. Thus, DDFI determines which evidence is shared, whereas DCM determines how it is adapted, coupling the two representations without forcing identical responses.

\subsection{Physical Consistency Learning}
At the output level, feature-space coupling does not guarantee agreement between the two estimates: the Stokes components computed from $\hat{\mathbf{I}}$ may differ from $\tilde{s}_1$ and $\tilde{s}_2$ even when both outputs are close to their references. PCL therefore uses \eqref{eq:polarimetric_relations} as a supervisory bridge. Applying it to $\hat{\mathbf{I}}$ gives $\hat{s}_0$, $\hat{s}_1$, and $\hat{s}_2$, and the first constraint aligns the two estimates of the difference components:
\begin{equation}
\mathcal{L}_{\mathrm{cons}}
:=
\left\lVert\hat{s}_{1}-\tilde{s}_{1}\right\rVert_{1}
+
\left\lVert\hat{s}_{2}-\tilde{s}_{2}\right\rVert_{1}.
\end{equation}
Component agreement does not account for AoLP periodicity or the feasible range of DoLP, so we constrain these properties separately.

According to \eqref{eq:polarimetric_relations}, the direction of the Stokes vector $(s_1,s_2)$ encodes twice the AoLP but is unreliable in weakly polarized regions. We therefore define a normalized direction and a DoLP-based confidence weight at position $i$:
\begin{equation}
\begin{aligned}
\mathbf{u}_{i}(s_1,s_2)
&:=
\frac{[s_{1,i},s_{2,i}]^{\mathrm{T}}}
{\sqrt{s_{1,i}^{2}+s_{2,i}^{2}+\epsilon}},\\
\omega_i
&:=
\operatorname{clip}
\left(\frac{p_i-\tau_l}{\tau_h-\tau_l},0,1
\right).
\end{aligned}
\end{equation}
Here, $p_i$ is the ground-truth DoLP, while $\tau_l$ and $\tau_h$ control the confidence weighting. With $\tilde{\mathbf{u}}_i=\mathbf{u}_i(\tilde{s}_1,\tilde{s}_2)$, $\mathbf{u}_i^{gt}=\mathbf{u}_i(s_1,s_2)$, and $c_i=\langle\tilde{\mathbf{u}}_i,\mathbf{u}_i^{gt}\rangle$, the angular fidelity loss is:
\begin{equation}
\mathcal{L}_{\mathrm{aop}}^{w}
:=
\frac{\sum_i\omega_i(1-c_i)}
{\sum_i\omega_i+\epsilon}.
\end{equation}
By emphasizing reliable DoLP regions, $\mathcal{L}_{\mathrm{aop}}^{w}$ aligns the predicted Stokes direction with the reference while avoiding excessive supervision where AoLP is poorly defined.

Meanwhile, a physically valid DoLP should not exceed one. Since
independently restored polarization images may violate this upper bound, we impose:

\begin{equation}
\mathcal{L}_{\mathrm{dop}}
:=
\frac{1}{N}\sum_i\operatorname{ReLU}(\hat{p}_i-1),
\end{equation}
where $N$ denotes the number of evaluated pixel-channel positions. Thus, $\mathcal{L}_{\mathrm{dop}}$ penalizes DoLP estimates above the physical upper bound without affecting feasible ones. Together, the component, direction, and feasibility terms couple the two outputs while respecting properties specific to linear-polarization estimation.

\subsection{Training Objective}
The frequency supervision and physical constraints complement direct data fidelity. We supervise the restored images and the predicted Stokes components with
\begin{equation}
\mathcal{L}_{\mathrm{img}}
=
\left\lVert\hat{\mathbf{I}}-\mathbf{I}\right\rVert_1,
\qquad
\mathcal{L}_{\mathrm{s12}}
=
\left\lVert\tilde{\mathbf{s}}_{\Delta}-\mathbf{s}_{\Delta}\right\rVert_1.
\end{equation}
Combining direct supervision, Laplacian frequency preservation, and output-space reconciliation gives the overall objective
\begin{equation}
\begin{aligned}
\mathcal{L}
=
&\mathcal{L}_{\mathrm{img}}
+\mathcal{L}_{\mathrm{s12}}
+\mathcal{L}_{\mathrm{freq}}
+\lambda_{1}\mathcal{L}_{\mathrm{cons}}
+\lambda_{2}\mathcal{L}_{\mathrm{aop}}^{w}
+\lambda_{3}\mathcal{L}_{\mathrm{dop}},
\end{aligned}
\end{equation}
where the weights balance direct reconstruction, Image--Stokes agreement, polarimetric constraints, and structural regularization. The resulting objective anchors each output to its reference while coupling the two predictions through the physical relations that they must share.

\section{Experiments}

\subsection{Experimental Setup}

\begin{table*}[!t]
  \caption{Quantitative comparison of task-specific polarimetric restoration, general multi-degradation restoration, and joint image--polarimetric restoration methods for compound-degraded CPFA observations.
  $p$ and $\theta$ denote DoLP and AoLP, respectively.
  Top-three results are highlighted in \textcolor{red}{red} (1st), \textcolor{blue}{blue} (2nd), and \textcolor{green!60!black}{green} (3rd). For input-matched comparison, $\ast$ denotes the variants that use the same observation-derived degraded Stokes cues as UniPolar. $\dagger$ indicates initialization with weights from denoising-and-demosaicking training on synthetic data~\cite{li2025demosaicking}.}
  \label{tab:compound_comparison}
  \centering
  \fontsize{7}{8}\selectfont
  \renewcommand{\arraystretch}{1.13}
  \setlength{\tabcolsep}{3.2pt}
  \arrayrulecolor{tblLine}

  \setlength{\fboxsep}{2pt}
  \begin{tabular}{l||ccc||ccc||ccc||ccc}
    \hline
    \textbf{Method}
    & \multicolumn{3}{c||}{\textit{Normal}}
    & \multicolumn{3}{c||}{\textit{Low-light}}
    & \multicolumn{3}{c||}{\textit{Haze}}
    & \multicolumn{3}{c}{\textit{Average}} \\
    \cline{2-13}
    & \textit{Mean} $\uparrow$ & $p$ $\uparrow$ & $\theta$ $\downarrow$
    & \textit{Mean} $\uparrow$ & $p$ $\uparrow$ & $\theta$ $\downarrow$
    & \textit{Mean} $\uparrow$ & $p$ $\uparrow$ & $\theta$ $\downarrow$
    & \textit{Mean} $\uparrow$ & $p$ $\uparrow$ & $\theta$ $\downarrow$ \\
    \hline
    \hline

    \multicolumn{13}{l}{\textit{\textbf{\textcolor{gray}{Task-Specific Polarimetric Methods}}}} \\
    \hline

    TCPDNet~\cite{nguyen2022two}
    & 36.790/0.930 & 27.721/0.587 & 19.152
    & 30.836/0.828 & 18.927/0.437 & 32.659
    & 14.190/0.499 & 16.534/0.291 & 29.391
    & 27.272/0.752 & 21.061/0.438 & 27.067 \\

    PIDSR~\cite{zhou2025pidsr}
    & \textcolor{green!60!black}{37.924}/0.939
    & \textcolor{green!60!black}{28.194}/0.602
    & 19.969
    & 32.308/0.881 & 20.431/0.499 & 27.696
    & 15.272/0.461 & 13.691/0.222 & 30.039
    & 28.501/0.760 & 20.772/0.441 & 25.901 \\

    CPDDNet~\cite{rahman2025polarization}
    & 36.085/0.921 & 26.930/0.554 & 22.087
    & 29.480/0.795 & 20.170/0.395 & 42.452
    & 14.324/0.507 & 16.614/0.299 & 28.080
    & 26.630/0.741 & 21.238/0.416 & 30.873 \\
    \hline
    \hline

    \multicolumn{13}{l}{\textit{\textbf{\textcolor{gray}{General Multi-Degradation Image Restoration Methods}}}} \\
    \hline

    AirNet~\cite{Li2022AllInOneIR}
    & 26.933/0.874 & 24.105/0.538 & 22.692
    & 30.962/0.883 & 21.066/0.544 & 22.593
    & 14.930/0.600 & 17.801/0.355 & 25.787
    & 24.275/0.786 & 20.991/0.479 & 23.691 \\


    OneRestore~\cite{Guo2024OneRestore}
    & 31.761/0.899 & 18.251/0.341 & 34.812
    & 13.265/0.482 & 10.089/0.250 & 48.251
    & 14.862/0.563 & 16.072/0.275 & 45.580
    & 19.963/0.648 & 14.804/0.289 & 42.881 \\


    AdaIR~\cite{Cui2025AdaIR}
    & 35.163/0.929
    & 26.752/0.594
    & 19.645
    & 33.877/0.926
    & 21.654/0.566
    & 24.782
    & 17.302/0.680
    & 17.915/0.382
    & 24.931
    & 28.780/0.845
    & 22.107/0.514
    & 23.120 \\


    AnyIR~\cite{Ren2026AnyIR}
    & 34.574/0.917
    & 26.330/0.571
    & 20.115
    & 33.034/0.912
    & 21.541/0.535
    & 25.740
    & 16.981/0.650
    & 18.140/0.364
    & 25.832
    & 28.196/0.826
    & 22.004/0.490
    & 23.896 \\


    PromptIR~\cite{Potlapalli2023PromptIRPF}
    & 34.759/0.921
    & 27.033/0.601
    & 18.781
    & 34.204/0.925
    & 21.423/0.554
    & 22.845
    & 17.767/0.688
    & 18.131/0.396
    & 25.723
    & 28.910/0.845
    & 22.196/0.517
    & 22.450 \\

    PromptIR$\ast$~\cite{Potlapalli2023PromptIRPF}
    & 35.493/0.934
    & 27.830/\textcolor{green!60!black}{0.621}
    & 17.597
    & 34.057/0.929
    & \textcolor{green!60!black}{23.645}/0.582
    & 20.772
    & 17.467/\textcolor{green!60!black}{0.704}
    & 17.566/\textcolor{green!60!black}{0.443}
    & 22.106
    & 29.006/0.856
    & 23.014/0.549
    & 20.158 \\

    DFPIR~\cite{Tian2025DFPIR}
    & 36.536/0.940
    & 26.399/0.563
    & 24.092
    & 33.041/\textcolor{green!60!black}{0.930}
    & 22.109/0.569
    & 22.733
    & 17.425/0.653
    & 17.104/0.356
    & 25.389
    & 29.001/0.841
    & 21.871/0.496
    & 24.071 \\

    DFPIR$\ast$~\cite{Tian2025DFPIR}
    & 36.477/\textcolor{green!60!black}{0.941}
    & 26.923/0.609
    & \textcolor{green!60!black}{17.276}
    & \textcolor{green!60!black}{34.506}/\textcolor{blue}{0.933}
    & \textcolor{red}{24.155}/\textcolor{green!60!black}{0.630}
    & \textcolor{green!60!black}{18.141}
    & \textcolor{green!60!black}{18.093}/0.698
    & \textcolor{green!60!black}{18.210}/0.415
    & \textcolor{green!60!black}{19.710}
    & \textcolor{green!60!black}{29.692}/\textcolor{green!60!black}{0.857}
    & \textcolor{green!60!black}{23.096}/\textcolor{green!60!black}{0.552}
    & \textcolor{green!60!black}{18.376} \\

    \hline
    \hline

    \multicolumn{13}{l}{\textit{\textbf{\textcolor{gray}{Joint Image–Polarimetric Restoration Method }}}} \\
    \hline

    \rowcolor{mygray}
    \textbf{Ours}
    & \textcolor{blue}{39.661}/\textcolor{blue}{0.950}
    & \textcolor{blue}{28.663}/\textcolor{blue}{0.666}
    & \textcolor{blue}{15.806}
    & \textcolor{blue}{34.704}/0.922
    & \textcolor{blue}{23.745}/\textcolor{blue}{0.652}
    & \textcolor{blue}{17.058}
    & \textcolor{blue}{20.200}/\textcolor{blue}{0.754}
    & \textcolor{blue}{19.881}/\textcolor{blue}{0.505}
    & \textcolor{blue}{17.636}
    & \textcolor{blue}{31.522}/\textcolor{blue}{0.875}
    & \textcolor{blue}{24.097}/\textcolor{blue}{0.608}
    & \textcolor{blue}{16.833} \\

    \rowcolor{mygray}
    \textbf{Ours$\dagger$}
    & \textcolor{red}{41.399}/\textcolor{red}{0.962}
    & \textcolor{red}{29.722}/\textcolor{red}{0.694}
    & \textcolor{red}{14.512}
    & \textcolor{red}{36.069}/\textcolor{red}{0.940}
    & 22.449/\textcolor{red}{0.677}
    & \textcolor{red}{14.061}
    & \textcolor{red}{20.898}/\textcolor{red}{0.821}
    & \textcolor{red}{20.960}/\textcolor{red}{0.582}
    & \textcolor{red}{15.218}
    & \textcolor{red}{32.788}/\textcolor{red}{0.908}
    & \textcolor{red}{24.377}/\textcolor{red}{0.651}
    & \textcolor{red}{14.597} \\
    \hline
  \end{tabular}
\end{table*}

\begin{figure*}[!t]
\centering
\includegraphics[width=1\linewidth]{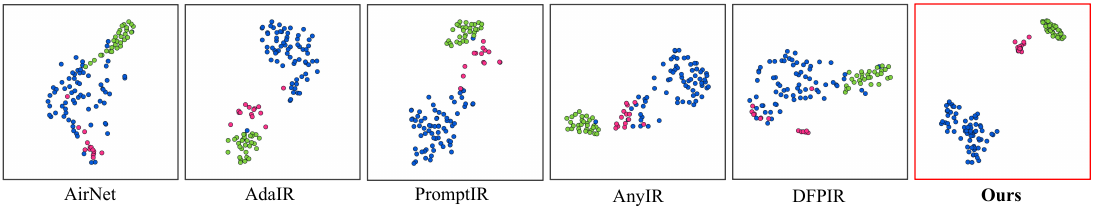}
\caption{t-SNE visualization of degradation representations. \textcolor{blue}{Blue}, \textcolor{green!60!black}{green} and \textcolor{magenta}{pink} denote normal, low-light, and haze conditions, respectively.}
\label{fig:tsne}
\end{figure*}

\subsubsection{Dataset}
We collect clean full-resolution four-angle images from publicly available polarimetric datasets~\cite{liu2023polarization,zhou2023polarization,wen2019convolutional,zhou2025pidsr,Morimatsu2021MonochromeAC} and synthesize haze from Sparse-PDM outdoor scenes because real hazy polarimetric data are scarce. The resulting dataset contains 528 training samples (359 normal, 100 low-light, and 69 haze) and 114 test samples (70 normal, 30 low-light, and 14 haze); the detailed per-source composition is provided in the supplementary material. Each sample comprises registered $I_0$, $I_{45}$, $I_{90}$, and $I_{135}$ images, to which we jointly apply downsampling, optional cropping, and normalization while preserving cross-angle alignment. During training, sparse CPFA mosaics are synthesized following the color--polarization sampling procedure of PIDSR~\cite{zhou2025pidsr}, together with the device- and environment-induced degradations considered here. We additionally evaluate joint denoising and demosaicking on PIDSR~\cite{zhou2025pidsr} and low-light enhancement on PLIE~\cite{zhou2023polarization}.

\subsubsection{Evaluation Metrics}
We evaluate the two objectives of joint image and polarimetric restoration. For image restoration, we compute peak signal-to-noise ratio (PSNR)~\cite{huynhthu2008scope} and structural similarity (SSIM)~\cite{wang2004image} for $I_0$, $I_{45}$, $I_{90}$, and $I_{135}$, and report their averages as the mean image score. For polarimetric restoration, $s_0=I_0+I_{90}=I_{45}+I_{135}$, $s_1=I_0-I_{90}$, and $s_2=I_{45}-I_{135}$ are computed independently for each RGB channel. We report PSNR/SSIM for $s_0/2$ and DoLP $p$ without first averaging the color channels. Image--Stokes agreement is evaluated by the MAE between the directly predicted and image-derived $s_1/s_2$ components.

AoLP is computed channel-wise as $\theta=\frac{1}{2}\operatorname{atan2}(s_2,s_1)$. Given the prediction--reference difference $\Delta\theta=\hat{\theta}-\theta$, we report the periodic angular error
\begin{equation}
e_{\theta}=\frac{180}{\pi}\left|\frac{1}{2}\operatorname{atan2}
\left(\sin(2\Delta\theta),\cos(2\Delta\theta)\right)\right|,
\end{equation}
averaged over all pixels and RGB channels. This formulation measures the shortest difference between two polarization directions while respecting the $\pi$-periodicity of AoLP, thereby avoiding artificially large errors for physically close directions near the interval boundary.

\subsubsection{Implementation Details}
UniPolar is implemented in PyTorch with a Restormer backbone~\cite{zamir2022restormer}. The input contains 12-channel sparse CPFA measurements, 12-channel masks, and 6-channel degraded $s_1/s_2$ cues computed from mask-aware locally filled observations. The model outputs four polarized images and Stokes $s_1/s_2$. Learnable scales $\alpha_{m,\mathrm{conv}}^{l}$ and $\alpha_{m,\mathrm{FiLM}}^{l}$ are both initialized to $0.01$. We set $(\lambda_1,\lambda_2,\lambda_3)=(0.2,0.03,0.01)$. The confidence parameters are $\tau_l=0.05$ and $\tau_h=0.20$. Input images are cropped into $256\times256$ patches for training. The model is trained for 500 epochs using AdamW with a learning rate of $1\times10^{-4}$ and a batch size of 2 on an NVIDIA RTX 4090. The degraded Stokes cues provide a coarse component prior without using clean ground truth. The base feature dimension is 48, with block depths $[4,6,6,8]$, attention heads $[1,2,4,8]$, and 4 refinement blocks. Each modulation block adopts a bottleneck ratio of 2. Normal, low-light, and haze conditions are encoded as 0, 1, and 2, respectively, with the task-matched condition provided during both training and inference. The frequency-loss weights are set to $\lambda_{f0}=0.03$ and $\lambda_{f12}=0.05$. The loss applies channel-wise $3\times3$ Laplacian operators to constrain the response magnitudes of the image-derived $s_0$ and predicted $s_1/s_2$. Its Stokes-side term and $\mathcal{L}_{\mathrm{aop}}^{w}$ are linearly warmed up during the first 10 epochs. All images are normalized to $[0,1]$ before training.

\subsection{Comparison with State-of-the-Art}
\noindent \textbf{Comparison under Compound Degradations.}
Table~\ref{tab:compound_comparison} compares task-specific polarimetric methods (TCPDNet~\cite{nguyen2022two}, PIDSR~\cite{zhou2025pidsr}, and CPDDNet~\cite{rahman2025polarization}) with general multi-degradation methods (AirNet~\cite{Li2022AllInOneIR}, OneRestore~\cite{Guo2024OneRestore}, AdaIR~\cite{Cui2025AdaIR}, AnyIR~\cite{Ren2026AnyIR}, PromptIR~\cite{Potlapalli2023PromptIRPF}, and DFPIR~\cite{Tian2025DFPIR}). All methods use identical data pairs and splits. Polarimetric baselines retain official CPFA inputs and are retrained per degradation; multi-degradation baselines are jointly retrained with $[\mathbf{X},\mathbf{M}]$, changing only their input/output projections. DFPIR and UniPolar both use the known task-matched degradation category during training and inference. For a fair comparison, we additionally provide PromptIR and DFPIR with the same observation-derived degraded $s_1/s_2$ cues used by UniPolar. 

As shown in Table~\ref{tab:compound_comparison}, UniPolar achieves the best average image and DoLP PSNRs and the lowest average AoLP error among the compared baselines, while maintaining competitive performance across normal, low-light, and haze conditions without additional pretraining. Although task-specific polarimetric methods are retrained for each degradation condition, their performance remains limited under compound degradations, particularly haze, where UniPolar improves the best competing image and DoLP PSNRs by 4.928~dB and 3.267~dB, respectively, while reducing the AoLP error by 10.444$^\circ$. These gains highlight the limitations of addressing device- and environment-induced degradations separately. Compared with the input-matched DFPIR$^\ast$, UniPolar improves the average image PSNR/SSIM by 1.830~dB/0.018 and the DoLP PSNR/SSIM by 1.001~dB/0.056, while reducing the AoLP error by 1.543$^\circ$. The advantage is particularly evident under haze, with gains of 2.107~dB in image PSNR and 1.671~dB in DoLP PSNR, alongside a 2.074$^\circ$ reduction in AoLP error. Under low-light conditions, DFPIR$^\ast$ achieves higher image SSIM and DoLP PSNR, whereas UniPolar obtains higher image PSNR and DoLP SSIM, as well as lower AoLP error. With prior initialization, UniPolar further improves the average image and DoLP scores to 32.788/0.908 and 24.377/0.651, respectively, while reducing AoLP error to 14.597$^\circ$. These results suggest that general multi-degradation methods, despite their ability to restore image appearance, struggle to preserve the cross-angle polarization relationships essential for reliable polarimetric recovery, highlighting the benefit of physically consistent Image--Stokes coupling.


\begin{table}[!t]
  \caption{Comparison with task-specific polarimetric restoration methods under individual device-based and environment-based degradation settings. $p$ and $\theta$ denote DoLP and AoLP, respectively. Top-three results are highlighted in \textcolor{red}{red} (1st), \textcolor{blue}{blue} (2nd), and \textcolor{green!60!black}{green} (3rd).}
  \label{tab:polar_comparison}
  \centering
  \fontsize{7}{8}\selectfont
  \renewcommand{\arraystretch}{1.1}
  \setlength{\tabcolsep}{3.3pt}
  \setlength{\arrayrulewidth}{0.55pt}
  \arrayrulecolor{tblLine}

  \setlength{\fboxsep}{2pt}
  \begin{tabular}{c|l||cccc}
    \hline
    \multicolumn{2}{c||}{\textbf{Method}}
    & \textit{Mean} $\uparrow$ & \textit{$s_0$} $\uparrow$ & $p$ $\uparrow$ & $\theta$ $\downarrow$ \\
    \hline
    \hline

    \multicolumn{6}{l}{\textit{\textbf{\textcolor{gray}{Low-Light Enhancement}}}} \\
    \hline

    \multirow{2}{*}{\rotatebox{90}{\textit{PLIE}}}
    & PLIE~\cite{zhou2023polarization}  & - & \textcolor{red}{39.1900}/\textcolor{red}{0.9770} & \textcolor{red}{27.1500}/\textcolor{red}{0.7650} & \textcolor{blue}{15.9859} \\
    & \cellcolor{mygray}\textbf{Ours}
    & \cellcolor{mygray}\textcolor{red}{36.0689}/\textcolor{red}{0.9401}
    & \cellcolor{mygray}\textcolor{blue}{37.0561}/\textcolor{blue}{0.9625}
    & \cellcolor{mygray}\textcolor{blue}{22.4486}/\textcolor{blue}{0.6765}
    & \cellcolor{mygray}\textcolor{red}{14.0609} \\
    \hline
    \hline

    \multicolumn{6}{l}{\textit{\textbf{\textcolor{gray}{Denoising and Demosaicking}}}} \\
    \hline

    \multirow{5}{*}{\rotatebox{90}{\textit{PIDSR}}}
    & Polanalyser~\cite{maeda2019polanalyser} & 32.1525/0.8971 & 33.2800/0.9138 & 26.6800/0.7164 & 17.8666 \\
    & IGRI2~\cite{Morimatsu2021MonochromeAC} & 34.4200/0.9361 & 35.5000/0.9450 & 27.7800/0.7486 & 16.5830 \\
    & TCPDNet~\cite{nguyen2022two} & \textcolor{green!60!black}{37.6425}/\textcolor{green!60!black}{0.9602} & \textcolor{green!60!black}{38.6500}/\textcolor{green!60!black}{0.9669} & \textcolor{green!60!black}{32.2600}/\textcolor{green!60!black}{0.8262} & \textcolor{green!60!black}{13.1881} \\
    & PIDSR~\cite{zhou2025pidsr} & \textcolor{blue}{39.0500}/\textcolor{blue}{0.9722} & \textcolor{blue}{40.2400}/\textcolor{blue}{0.9780} & \textcolor{blue}{33.3300}/\textcolor{blue}{0.8447} & \textcolor{blue}{12.2383} \\
    & \cellcolor{mygray}\textbf{Ours}
    & \cellcolor{mygray}\textcolor{red}{44.7447}/\textcolor{red}{0.9852}
    & \cellcolor{mygray}\textcolor{red}{45.8997}/\textcolor{red}{0.9883}
    & \cellcolor{mygray}\textcolor{red}{33.4220}/\textcolor{red}{0.8612}
    & \cellcolor{mygray}\textcolor{red}{11.4149} \\
    \hline
  \end{tabular}
\end{table}

\begin{table}[!t]
  \caption{Comparison with task-specific methods for joint denoising and demosaicking on Synthetic MQ and Synthetic PIDSR~\cite{li2025demosaicking}. $p$ and $\theta$ denote DoLP and AoLP, respectively. Top-three results are highlighted in \textcolor{red}{red} (1st), \textcolor{blue}{blue} (2nd), and \textcolor{green!60!black}{green} (3rd).}
  \label{tab:supp_synthetic_individual}
  \centering
  \fontsize{7}{8}\selectfont
  \renewcommand{\arraystretch}{1.12}
  \setlength{\tabcolsep}{3.3pt}
  \setlength{\arrayrulewidth}{0.55pt}
  \arrayrulecolor{tblLine}
  \begin{tabular}{c|l||cccc}
    \hline
    \multicolumn{2}{c||}{\textbf{Method}}
    & \textit{Mean} $\uparrow$ & \textit{$s_0$} $\uparrow$ & $p$ $\uparrow$ & $\theta$ $\downarrow$ \\
    \hline
    \hline
    \multirow{8}{*}{\rotatebox{90}{\textit{Synthetic MQ}}}
    & Polanalyser~\cite{maeda2019polanalyser} & 38.2149/0.9653 & 35.2560/0.9541 & 32.5532/0.8387 & 14.8220 \\
    & NLCSR~\cite{luo2024learning} & 36.8877/0.9523 & 33.5610/0.9310 & 34.0964/0.8557 & 11.1297 \\
    & CPDNet~\cite{wen2019convolutional} & 39.4374/0.9700 & 36.6062/0.9631 & 32.6586/0.8465 & 12.5138 \\
    & TCPDNet~\cite{nguyen2022two} & \textcolor{green!60!black}{42.5602}/0.9815 & 39.7694/0.9765 & 36.0613/0.8921 & 10.1765 \\
    & DCPM~\cite{li2025demosaicking} & 42.0221/0.9807 & \textcolor{green!60!black}{40.0263}/0.9759 & 36.1496/\textcolor{green!60!black}{0.8958} & 11.7376 \\
    & PIDSR~\cite{zhou2025pidsr} & 42.4990/\textcolor{green!60!black}{0.9822} & \textcolor{blue}{40.6269}/\textcolor{blue}{0.9794} & \textcolor{green!60!black}{36.5413}/0.8950 & \textcolor{blue}{9.6463} \\
    & PUGDiff~\cite{li2026polarization} & \textcolor{blue}{42.6794}/\textcolor{blue}{0.9827} & 39.4785/\textcolor{green!60!black}{0.9768} & \textcolor{red}{37.4700}/\textcolor{blue}{0.9059} & \textcolor{red}{9.2029} \\
    & \cellcolor{mygray}\textbf{Ours}
    & \cellcolor{mygray}\textcolor{red}{43.8110}/\textcolor{red}{0.9836}
    & \cellcolor{mygray}\textcolor{red}{44.7306}/\textcolor{red}{0.9865}
    & \cellcolor{mygray}\textcolor{blue}{37.1671}/\textcolor{red}{0.9104}
    & \cellcolor{mygray}\textcolor{green!60!black}{9.8178} \\
    \hline
    \hline
    \multirow{8}{*}{\rotatebox{90}{\textit{Synthetic PIDSR}}}
    & Polanalyser~\cite{maeda2019polanalyser} & 41.0285/0.9655 & 39.4167/0.9621 & 35.8349/0.8347 & 17.9892 \\
    & NLCSR~\cite{luo2024learning} & 40.0285/0.9601 & 38.5526/0.9569 & 36.0323/0.8775 & 16.0549 \\
    & CPDNet~\cite{wen2019convolutional} & 41.4389/0.9694 & 39.6154/0.9639 & 37.0999/0.8529 & \textcolor{blue}{14.9234} \\
    & TCPDNet~\cite{nguyen2022two} & 43.1434/0.9785 & 41.2190/0.9734 & 39.1320/0.8883 & 15.5998 \\
    & DCPM~\cite{li2025demosaicking} & \textcolor{green!60!black}{43.4987}/\textcolor{green!60!black}{0.9804} & \textcolor{green!60!black}{41.7152}/\textcolor{green!60!black}{0.9750} & 38.4340/0.8791 & \textcolor{green!60!black}{15.5325} \\
    & PIDSR~\cite{zhou2025pidsr} & 43.0540/0.9792 & 41.2950/0.9741 & \textcolor{blue}{39.8428}/\textcolor{blue}{0.9015} & 16.4804 \\
    & PUGDiff~\cite{li2026polarization} & \textcolor{blue}{44.2325}/\textcolor{blue}{0.9815} & \textcolor{blue}{41.9823}/\textcolor{blue}{0.9753} & \textcolor{red}{40.6696}/\textcolor{red}{0.9107} & 17.0324 \\
    & \cellcolor{mygray}\textbf{Ours}
    & \cellcolor{mygray}\textcolor{red}{45.6767}/\textcolor{red}{0.9876}
    & \cellcolor{mygray}\textcolor{red}{46.7585}/\textcolor{red}{0.9893}
    & \cellcolor{mygray}\textcolor{green!60!black}{39.4517}/\textcolor{green!60!black}{0.8942}
    & \cellcolor{mygray}\textcolor{red}{11.8615} \\
    \hline
  \end{tabular}
\end{table}

Figure~\ref{fig:tsne} visualizes the learned degradation representations. Compared with DFPIR and the other general multi-degradation baselines, UniPolar forms more compact intra-class clusters and clearer inter-class separation, suggesting more discriminative degradation representations.

\noindent \textbf{Comparison under Individual Degradations.}
Table~\ref{tab:polar_comparison} evaluates low-light enhancement on PLIE against PLIE~\cite{zhou2023polarization} and joint denoising and demosaicking on PIDSR against Polanalyser~\cite{maeda2019polanalyser}, IGRI2~\cite{Morimatsu2021MonochromeAC}, TCPDNet~\cite{nguyen2022two}, and PIDSR~\cite{zhou2025pidsr}. Table~\ref{tab:supp_synthetic_individual} additionally compares NLCSR~\cite{luo2024learning}, CPDNet~\cite{wen2019convolutional}, DCPM~\cite{li2025demosaicking}, and PUGDiff~\cite{li2026polarization} on Synthetic MQ and Synthetic PIDSR. All methods use matched degraded observations, clean references, and evaluation code; task-specific baselines follow their official protocols, while UniPolar derives $\mathbf{X}$, $\mathbf{M}$, and degraded $s_1/s_2$ cues from the same observation with the task-matched condition.

Under these matched settings, as shown in Table~\ref{tab:polar_comparison}, the mean image score on PLIE is unavailable because PLIE neither takes the four angle-resolved images as input nor restores them. Although PLIE achieves higher $s_0$ and DoLP PSNR, UniPolar obtains a higher DoLP SSIM of 0.6765 and reduces the AoLP error from 15.9859$^\circ$ to 14.0609$^\circ$. On PIDSR, UniPolar ranks first across all reported metrics, including a mean image score of 44.7447/0.9852 and an AoLP error of 11.4149$^\circ$. Table~\ref{tab:supp_synthetic_individual} further evaluates joint polarimetric denoising and demosaicking. On Synthetic MQ, UniPolar achieves the best mean image and $s_0$ scores, improving mean image PSNR over PUGDiff~\cite{li2026polarization} by 1.1316~dB. It also obtains the highest DoLP SSIM, although its AoLP error remains slightly higher than those of PUGDiff and PIDSR. On Synthetic PIDSR~\cite{zhou2025pidsr}, UniPolar again ranks first in mean image and $s_0$ reconstruction, with PSNR gains of 1.4442 and 4.7762~dB over PUGDiff, respectively. It also yields the lowest AoLP error, whereas PUGDiff retains the best DoLP result. Overall, these comparisons indicate that UniPolar provides a favorable balance between image fidelity, intensity reconstruction, and angular accuracy under individual degradations.

\begin{table}[!t]
  \caption{Component Ablation Study.}
  \label{tab:component_ablation}
  \centering
  \fontsize{7}{8}\selectfont
  \renewcommand{\arraystretch}{1.08}
  \setlength{\tabcolsep}{4.7pt}
  \setlength{\arrayrulewidth}{0.55pt}
  \arrayrulecolor{tblLine}
  \setlength{\fboxsep}{2pt}
  \begin{tabular}{cccc||ccc}
    \hline
    \textbf{DD} & \textbf{DDFI} & \textbf{DCM} & \textbf{PCL}
    & \textit{Mean} $\uparrow$ & $p$ $\uparrow$ & $\theta$ $\downarrow$ \\
    \hline
    \hline
          &       &       &       & 27.7079/0.7987 & 22.0222/0.4899 & 24.6531 \\
    $\checkmark$ &       &       &       & 28.9621/0.8338 & 23.1951/0.5281 & 20.9951 \\
    $\checkmark$ & $\checkmark$ &       &       & 29.4692/0.8438 & 24.0186/0.5588 & 19.4735 \\
    $\checkmark$ & $\checkmark$ & $\checkmark$ &       & 30.8831/0.8590 & 23.8901/0.5617 & 19.4576\\
    \rowcolor{mygray}
    $\checkmark$ & $\checkmark$ & $\checkmark$ & $\checkmark$
    & \textcolor{red}{31.5219}/\textcolor{red}{0.8752}
    & \textcolor{red}{24.0965}/\textcolor{red}{0.6076}
    & \textcolor{red}{16.8334} \\
    \hline
  \end{tabular}
\end{table}

\subsection{Ablation Study and Model Analysis}

\subsubsection{Ablation Study}

Table~\ref{tab:component_ablation} evaluates the components through progressive integration under compound degradations. Introducing DD improves the mean image score from 27.7079/0.7987 to 28.9621/0.8338 and the DoLP score from 22.0222/0.4899 to 23.1951/0.5281, while reducing AoLP error by 3.6580$^\circ$. These gains suggest that separate image and Stokes representations help address the distinct effects of degradations on visual structures and polarization differences. DDFI further improves mean image and DoLP PSNRs by 0.5071~dB and 0.8235~dB, respectively, while reducing AoLP error by 1.5216$^\circ$, supporting frequency-selective interaction against environmental variations and device-induced artifacts. DCM achieves the largest single-stage image PSNR gain of 1.4139~dB, reaching 30.8831~dB. However, DoLP SSIM increases to 0.5617 despite a 0.1285~dB decrease in DoLP PSNR, suggesting that condition-adaptive modulation primarily benefits visual restoration without uniformly improving polarization magnitude. PCL further improves DoLP SSIM by 0.0459 and reduces AoLP error by 2.6242$^\circ$, raising the mean image score to 31.5219/0.8752. This highlights the role of explicit Image--Stokes consistency in mitigating residual polarimetric errors beyond feature adaptation. Overall, the full model improves mean image PSNR/SSIM by 3.8140~dB/0.0765 and DoLP PSNR/SSIM by 2.0743~dB/0.1177, with a 7.8197$^\circ$ reduction in AoLP error. These results support the complementary roles of degradation-aware Image--Stokes interaction and physical consistency learning in jointly improving visual fidelity and polarimetric reliability.

\subsubsection{Analysis of Dual-Domain Frequency Interaction}
\begin{figure}[!t]
\centering
\includegraphics[width=1\linewidth]{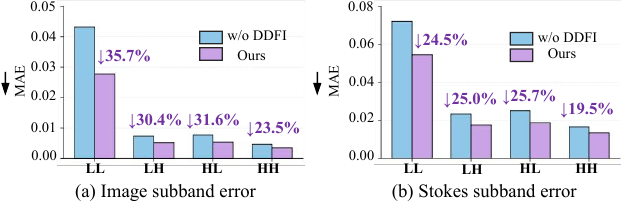}
\caption{Frequency-band error analysis.}
\label{fig:freq}
\end{figure}

Table~\ref{tab:freq_ablation} demonstrates the complementary roles of frequency interactions under compound degradations. The low-frequency branch improves mean-image and DoLP PSNRs by 0.1545 and 0.5611~dB over high-frequency interaction alone, suggesting better handling of illumination and scattering variations. In contrast, high-frequency interaction yields slightly higher mean-image SSIM by preserving structures affected by CPFA sampling and noise. Combining both reduces AoLP error to 17.3958$^\circ$, outperforming the individual branches by 0.6974$^\circ$ and 0.8882$^\circ$. Adding $\mathcal{L}_{\mathrm{freq}}$ further improves mean-image and DoLP PSNRs by 0.6412 and 0.6497~dB, respectively, while reducing AoLP error by 0.5624$^\circ$. The subband MAE reductions in Fig.~\ref{fig:freq} further support frequency-selective Image--Stokes interaction in mitigating compound-degradation errors.

\begin{table}[!t]
  \caption{Analysis of Dual-Domain Frequency Interaction.}
  \label{tab:freq_ablation}
  \centering
  \fontsize{7}{8}\selectfont
  \renewcommand{\arraystretch}{1.08}
  \setlength{\tabcolsep}{4pt}
  \setlength{\arrayrulewidth}{0.55pt}
  \arrayrulecolor{tblLine}

  \setlength{\fboxsep}{2pt}
  \begin{tabular}{ccc||ccc}
    \hline
    \textbf{Low-freq.} & \textbf{High-freq.} & $\boldsymbol{\mathcal{L}_{\mathrm{freq}}}$
    & \textit{Mean} $\uparrow$ & $p$ $\uparrow$ & $\theta$ $\downarrow$ \\
    \hline
    \hline
    $\checkmark$ &  &  & 30.7282/0.8552 & 23.4753/0.5693 & 18.0932 \\
    & $\checkmark$ &  & 30.5737/0.8565 & 22.9142/0.5677 & 18.2840 \\
    $\checkmark$ & $\checkmark$ &  & 30.8807/0.8603 & 23.4468/0.5800 & 17.3958 \\
    \rowcolor{mygray}
    $\checkmark$ & $\checkmark$ & $\checkmark$
    & \textcolor{red}{31.5219}/\textcolor{red}{0.8752}
    & \textcolor{red}{24.0965}/\textcolor{red}{0.6076}
    & \textcolor{red}{16.8334} \\
    \hline
  \end{tabular}
\end{table}

\begin{table}[!t]
\caption{Analysis of Degradation-Conditioned Modulation.}
\label{tab:modulation_ablation}
\centering
\fontsize{7}{8}\selectfont
\renewcommand{\arraystretch}{1.08}
\setlength{\tabcolsep}{9pt}
\setlength{\arrayrulewidth}{0.55pt}
\arrayrulecolor{tblLine}

\begin{tabular}{cc||ccc}
\hline
\textbf{Image} & \textbf{Stokes}
& \textit{Mean} $\uparrow$ & $p$ $\uparrow$ & $\theta$ $\downarrow$ \\
\hline
\hline

$\checkmark$ &
& 31.2725/0.8665 & 23.4205/0.5800 & 17.5793 \\

& $\checkmark$
& 30.6966/0.8563 & 23.4003/0.6306 & 18.1634 \\

\rowcolor{mygray}
$\checkmark$ & $\checkmark$
& \textcolor{red}{31.5219}/\textcolor{red}{0.8752}
    & \textcolor{red}{24.0965}/\textcolor{red}{0.6076}
    & \textcolor{red}{16.8334} \\
\hline

\end{tabular}
\end{table}

\subsubsection{Analysis of Degradation-Conditioned Modulation}
Table~\ref{tab:modulation_ablation} examines degradation-conditioned modulation across both streams. Image-stream modulation achieves 0.5759~dB higher mean-image PSNR and 0.5841$^\circ$ lower AoLP error than Stokes-stream modulation, whereas the latter yields the highest individual DoLP SSIM of 0.6306. Combining both streams improves mean-image PSNR/SSIM by 0.2494~dB/0.0087 and DoLP PSNR/SSIM by 0.6760~dB/0.0276, while reducing AoLP error by 0.7459$^\circ$ over image-only modulation. These results suggest that compound degradations require condition-adaptive, stream-specific corrections to address their distinct effects on image appearance and polarization information.

\subsubsection{Analysis of Physical Consistency Learning}
\begin{figure}[!t]
\centering
\includegraphics[width=1\linewidth]{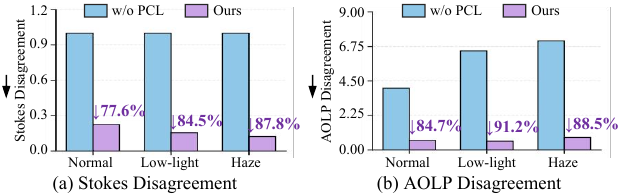}
\caption{Quantitative analysis of Image--Stokes physical consistency.}
\label{fig:pcl}
\end{figure}

Table~\ref{tab:pcl_ablation} evaluates the complementary physical constraints of PCL. Building on component consistency, $\mathcal{L}_{\mathrm{aop}}^{w}$ reduces AoLP error from 19.1311$^\circ$ to 17.3279$^\circ$, while $\mathcal{L}_{\mathrm{dop}}$ improves DoLP SSIM from 0.5799 to 0.5935 by penalizing infeasible polarization degrees. Combining all three constraints improves mean-image PSNR by 0.5285~dB and DoLP SSIM by 0.0277, while reducing AoLP error by 2.2977$^\circ$. Figure~\ref{fig:pcl} further shows 77.6--87.8\% lower Image--Stokes disagreement and 84.7--91.2\% lower AoLP disagreement across all conditions. These results highlight the need for explicit physical consistency, as improved image fidelity alone does not ensure reliable polarimetric recovery under compound degradations.

\begin{table}[!t]
  \caption{Analysis of Physical Consistency Learning.}
  \label{tab:pcl_ablation}
  \centering
  \fontsize{7}{8}\selectfont
  \renewcommand{\arraystretch}{1.08}
  \setlength{\tabcolsep}{5.7pt}
  \arrayrulecolor{tblLine}
  \setlength{\fboxsep}{2pt}

\begin{tabular}{ccc||ccc}
\hline
$\boldsymbol{\mathcal{L}_{\mathrm{cons}}}$ & $\boldsymbol{\mathcal{L}_{\mathrm{aop}}^{w}}$ & $\boldsymbol{\mathcal{L}_{\mathrm{dop}}}$
    & \textit{Mean} $\uparrow$ & $p$ $\uparrow$ & $\theta$ $\downarrow$ \\
\hline
\hline
$\checkmark$ &
&
& 30.9934/0.8677 & 23.9921/0.5799 & 19.1311 \\

$\checkmark$ & $\checkmark$ &
& 31.3422/0.8701 & 23.8743/0.5701 & 17.3279 \\

$\checkmark$ & & $\checkmark$
& 31.2095/0.8691 & 23.9910/0.5935 & 18.5869 \\

\rowcolor{mygray}
$\checkmark$ & $\checkmark$ & $\checkmark$
& \textcolor{red}{31.5219}/\textcolor{red}{0.8752}
    & \textcolor{red}{24.0965}/\textcolor{red}{0.6076}
    & \textcolor{red}{16.8334} \\
\hline
\end{tabular}
\end{table}

\section{Qualitative Results}

\begin{figure*}[!t]
\centering
\includegraphics[width=\linewidth]{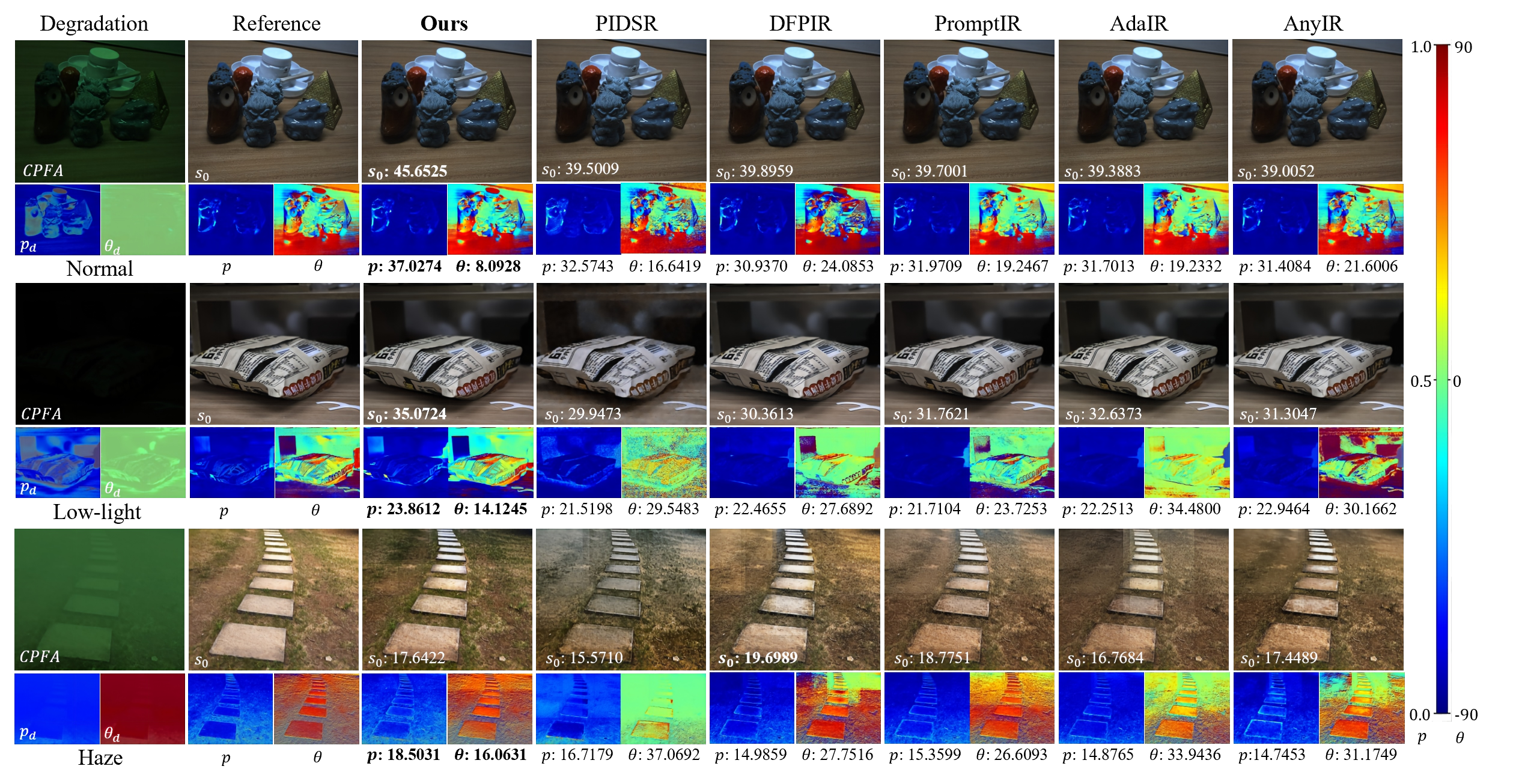}
\caption{Qualitative comparisons under normal, low-light, and haze conditions. For each scene, we show restored total intensity $s_0$, DoLP $p$, and AoLP $\theta$.}
\label{fig:vis_compare}
\end{figure*}

\begin{figure}[!t]
\centering
\includegraphics[width=\linewidth]{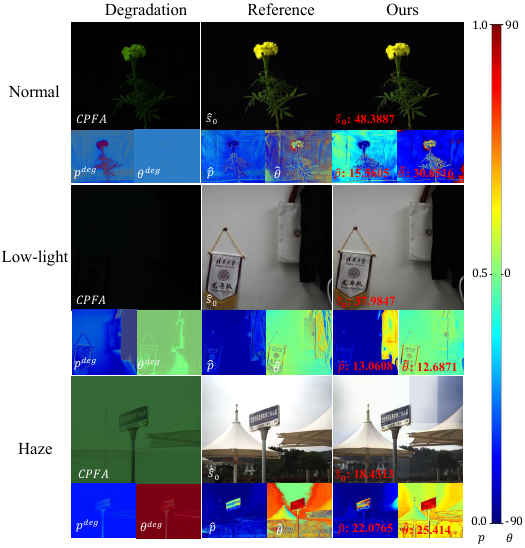}
\caption{Visualization of failure cases.}
\label{fig:vis_failure}
\end{figure}

\noindent\textbf{Qualitative Comparison.}
Figure~\ref{fig:vis_compare} compares UniPolar with task-specific polarimetric and general multi-degradation methods under normal, low-light, and haze conditions. We visualize restored total intensity $s_0$ together with DoLP and AoLP to assess both image quality and polarimetric reliability. Under normal conditions, most methods recover recognizable intensity structures, but competing results contain visible deviations in the DoLP and AoLP maps around object boundaries and material transitions. UniPolar preserves sharper structures and produces polarization distributions closer to the reference. Under low light, task-specific restoration leaves insufficient brightness, while several general multi-degradation methods improve overall appearance but blur local details or introduce angular errors. UniPolar restores foreground contrast while retaining coherent polarization patterns. Under haze, competing methods either leave residual scattering or recover visually plausible intensity images with inaccurate DoLP and AoLP. UniPolar suppresses the haze-induced veil while preserving scene boundaries and polarization variations.

\noindent\textbf{Analysis of Failure Cases.}
Figure~\ref{fig:vis_failure} shows representative failures under normal, low-light, and haze conditions. Under normal conditions, weak polarization in dark regions and thin structures amplifies small $s_1/s_2$ errors in the derived DoLP and AoLP. Under low light, severe signal attenuation reduces the reliability of sparse CPFA measurements and degraded Stokes cues, mainly affecting polarization magnitude recovery. Under haze, spatially varying attenuation and scattering-induced polarization cannot be fully represented by a single degradation code, leaving residual errors in both intensity and Stokes responses. Overall, errors are more pronounced in regions with weak polarization, low signal-to-noise ratios, or spatially varying scattering effects.

\section{Conclusion}
\label{sec:conclusion}
We presented UniPolar for joint image and polarimetric restoration from compound-degraded CPFA observations. Its coupled Image--Stokes streams preserve complementary representations, while DDFI and DCM enable frequency-selective information exchange and degradation-conditioned adaptation. PCL further promotes physical consistency through component, angular, and feasibility constraints. Experiments across normal, low-light, and haze conditions demonstrate improvements in both image fidelity and polarimetric accuracy over representative task-specific and general multi-degradation restoration methods, particularly in AoLP estimation under challenging conditions. These results highlight the value of explicitly considering polarimetric reliability alongside visual fidelity in learning-based restoration.

\appendices

\section{Dataset Generation} 
This section details the construction of the compound-degradation dataset used to train and evaluate UniPolar. We first assemble aligned clean polarization-angle images, then reproduce CPFA acquisition through mosaic sampling and introduce sensor noise and haze to model coupled device- and environment-induced degradations.

\subsection{Dataset Composition} 
\begin{figure}[!t]
\centering
\includegraphics[width=1\linewidth]{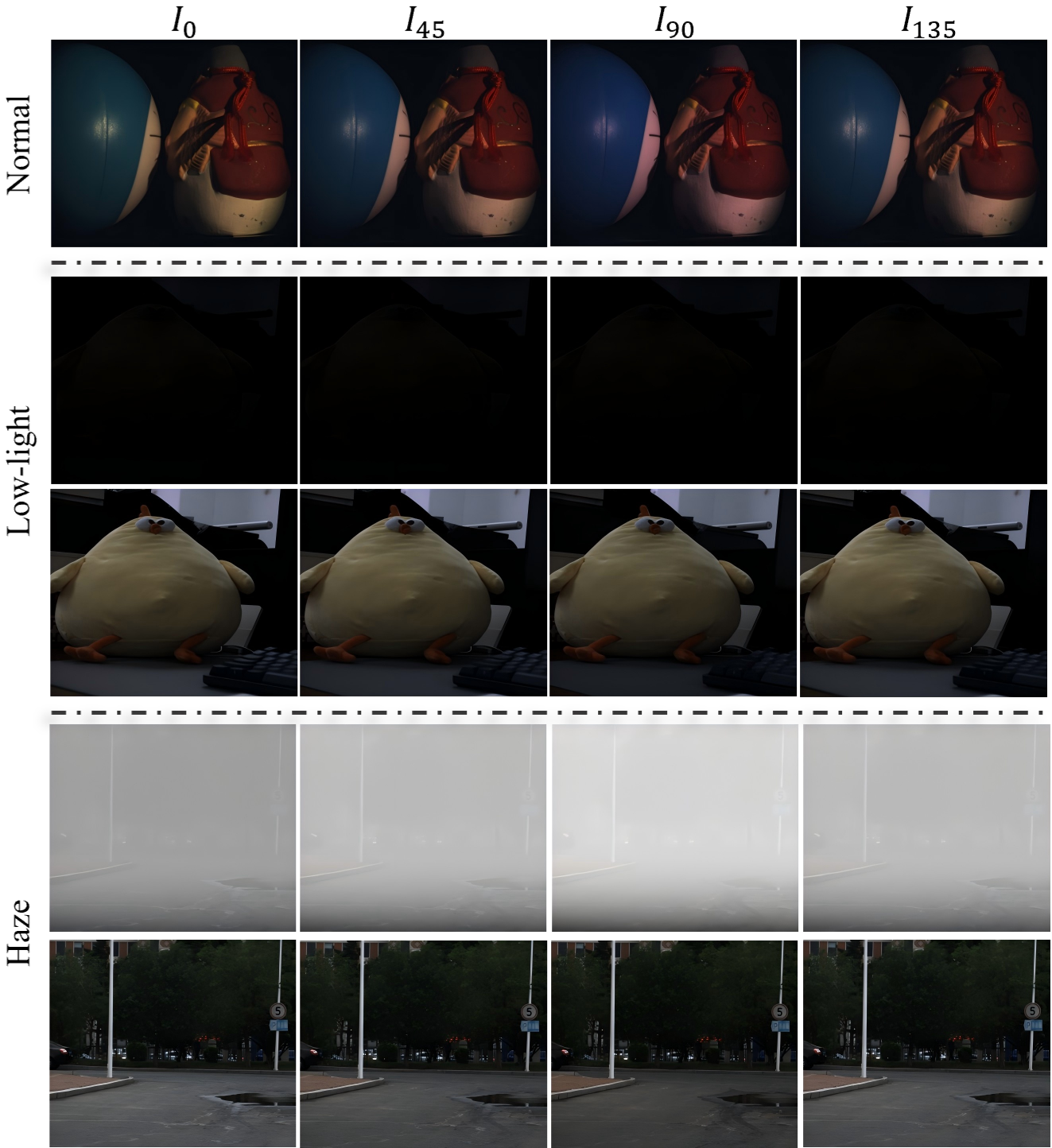}
\caption{Sample examples from the constructed dataset.}
\label{fig:data}
\end{figure}

As shown in Table~\ref{tab:dataset_composition_task}, we collect clean full-resolution polarized images with different polarizer angles from publicly available datasets~\cite{liu2023polarization,zhou2023polarization,wen2019convolutional,zhou2025pidsr,Morimatsu2021MonochromeAC}. Each sample contains four registered images, denoted by $I_0$, $I_{45}$, $I_{90}$, and $I_{135}$, which serve as reference targets for constructing the compound-degradation CPFA restoration dataset. Since publicly available hazy polarimetric images are particularly scarce, we synthesize the haze samples using outdoor scenes from Sparse-PDM~\cite{liu2023polarization}, whose scene content is suitable for haze simulation. The final dataset contains 642 samples. The training split consists of 359 normal, 100 low-light, and 69 haze samples, whereas the test split contains 70 normal, 30 low-light, and 14 haze samples, yielding 528 training samples and 114 test samples in total. Some examples are shown in Fig.~\ref{fig:data}.

To increase training diversity, we jointly apply the same downsampling and optional cropping operations to all four images, preserving pixel-wise alignment across polarizer angles. The augmented images are normalized to $[0,1]$ and stored as clean targets. This joint processing maintains cross-angle correspondence for reliable computation of Stokes parameters, DoLP, and AoLP.

\subsection{CPFA Mosaic Sampling}

To simulate the imaging process of a color-polarization filter array (CPFA), we convert the clean full-resolution polarized images with different polarizer angles into spatially multiplexed raw observations. In a CPFA sensor, each pixel measures only one color component at one polarizer angle. Therefore, the captured raw image is inherently sparse in both the color and polarization dimensions.

Given clean polarized images with different polarizer angles
$\{I_{0}, I_{45}, I_{90}, I_{135}\}$, we first decompose each image into RGB channels. Let $I_{\phi}^{c}$ denote the clean image acquired at polarizer angle $\phi \in \{0^\circ,45^\circ,90^\circ,135^\circ\}$ and color channel $c \in \{\mathrm{R},\mathrm{G},\mathrm{B}\}$. According to the CPFA sampling pattern, we generate binary masks $M_{\phi}^{c}$, where $M_{\phi}^{c}(\mathbf{x})=1$ indicates that pixel $\mathbf{x}$ observes color channel $c$ at polarizer angle $\phi$, and $0$ otherwise. The sparse CPFA observation is then obtained by

\begin{equation}
X_{\phi}^{c} = M_{\phi}^{c} \odot I_{\phi}^{c},
\end{equation}

where $\odot$ denotes element-wise multiplication. Stacking all sparse observations and masks gives the model inputs $\mathbf{X}=[X_{\phi}^{c}]_{\phi,c}$ and $\mathbf{M}=[M_{\phi}^{c}]_{\phi,c}$ used in the main paper. The raw CPFA mosaic combines these observations as

\begin{equation}
X_{\mathrm{CPFA}}(\mathbf{x})=M_{\phi(\mathbf{x})}^{c(\mathbf{x})}(\mathbf{x})\,I_{\phi(\mathbf{x})}^{c(\mathbf{x})}(\mathbf{x}),
\end{equation}

where $\phi(\mathbf{x})$ and $c(\mathbf{x})$ denote the polarizer angle and color channel sampled at pixel $\mathbf{x}$ according to the CPFA pattern.

\begin{table}[!t]
  \caption{Dataset composition and degradation condition.}
  \label{tab:dataset_composition_task}
  \centering
  \fontsize{8}{8}\selectfont
  \renewcommand{\arraystretch}{1.2}
  \setlength{\tabcolsep}{10pt}
  \setlength{\arrayrulewidth}{0.55pt}
  \arrayrulecolor{tblLine}
  \begin{tabular}{ll||ccc}
    \hline
    \textbf{Dataset} & \textbf{Task}
    & \textit{Train} & \textit{Test} & \textit{Total} \\
    \hline
    \hline
    CPDNet~\cite{wen2019convolutional} & Normal & 85 & 20 & 105 \\
    PIDSR~\cite{zhou2025pidsr} & Normal & 108 & 30 & 138 \\
    Sparse-PDM~\cite{liu2023polarization} & Normal & 126 & 20 & 146 \\
    Sparse-PDM~\cite{liu2023polarization} & Haze & 69 & 14 & 83 \\
    Tokyo~\cite{Morimatsu2021MonochromeAC} & Normal & 40 & -- & 40 \\
    PLIE~\cite{zhou2023polarization} & Low-light & 100 & 30 & 130 \\
    \hline
    \textbf{Total} & -- & \textbf{528} & \textbf{114} & \textbf{642} \\
    \hline
  \end{tabular}
\end{table}

In our implementation, we retain the sparse observations and their corresponding masks as separate channels. Specifically, the CPFA input contains 12 sparse observation channels and 12 binary mask channels for four polarizer angles and three color channels. This representation preserves the CPFA sampling process while indicating which pixels are observed and which require reconstruction. Mosaic sampling constitutes a device-based degradation because each polarized image is only partially observed. Restoring full-resolution $I_0$, $I_{45}$, $I_{90}$, and $I_{135}$ therefore requires recovering missing polarizer-angle and color samples from the spatially multiplexed measurements.

\subsection{Sensor Noise Simulation}

We use a heteroscedastic Gaussian approximation to Poisson--Gaussian noise to simulate device-based sensor corruption in CPFA polarimetric imaging.
Compared with additive white Gaussian noise, this model better reflects the combined effects of signal-dependent photon shot noise and signal-independent readout noise.

Given a normalized polarized image \(I\), the noisy observation is generated as
\begin{equation}
I_{\mathrm{noise}}
=
\operatorname{clip}
\left(
I + n_{\mathrm{shot}} + n_{\mathrm{read}},
0,1
\right),
\end{equation}
where the shot noise and read noise are modeled as
\begin{equation}
n_{\mathrm{shot}} \sim \mathcal{N}
\left(
0, \sigma^2 I
\right),
\qquad
n_{\mathrm{read}} \sim \mathcal{N}
\left(
0, (0.5\sigma)^2
\right).
\end{equation}
Here, \(\sigma\) controls the overall noise level. 
The variance of the signal-dependent term \(n_{\mathrm{shot}}\) increases with image intensity, while \(n_{\mathrm{read}}\) represents approximately intensity-independent sensor readout noise.

In our implementation, the noise level is selected from three settings:
\begin{equation}
\sigma \in \{0.01, 0.03, 0.05\},
\end{equation}
corresponding to low, medium, and high noise levels. 
After noise simulation, the noisy polarized images with different polarizer angles are converted into sparse CPFA observations using the fixed CPFA sampling pattern. 
This process provides a physically motivated approximation of device-based degradation caused by photon shot noise, sensor readout noise, and CPFA mosaic sampling.

\subsection{Haze Degradation}

We synthesize haze degradation on dense polarized images with different polarizer angles before training.
Unlike low-light degradation, the hazy images are precomputed for each polarizer angle.
CPFA mosaic sampling and sensor noise are not included in this offline haze synthesis stage, but are applied online during training.

Given clean polarized images with different polarizer angles
\(\mathbf{I}=\{I_{0}, I_{45}, I_{90}, I_{135}\}\),
we first estimate a relative inverse-depth map \(D\) using the pretrained Depth Anything model~\cite{Yang2024DepthAnything}. The depth predictor is applied to a representative RGB intensity image of the same scene, and the resulting scene-level depth prior is shared by all four polarizer angles to preserve cross-angle geometric consistency. Since Depth Anything provides relative rather than metric depth, we normalize \(D\) within each image and convert it into a depth-like map:
\begin{equation}
Z(\mathbf{x}) =
1-
\frac{D(\mathbf{x})-D_{\min}}
{D_{\max}-D_{\min}+\epsilon}.
\end{equation}
Here, $D_{\min}$ and $D_{\max}$ are the minimum and maximum of $D$, and $\epsilon$ is a small constant for numerical stability.
The transmission is then computed by the atmospheric scattering model:
\begin{equation}
t(\mathbf{x})=\exp(-\beta_{\mathrm{scat}} Z(\mathbf{x})),
\end{equation}
where \(\beta_{\mathrm{scat}}\) is the scattering coefficient.

Using the transmission map, we generate hazy polarized images with different polarizer angles using a physics-based polarimetric haze model.
The model accounts for attenuated scene radiance and partially polarized airlight, producing direction-dependent haze responses.
This produces dense hazy images
\(\mathbf{I}^{\mathrm{haze}}=\{I_{0}^{\mathrm{haze}}, I_{45}^{\mathrm{haze}}, I_{90}^{\mathrm{haze}}, I_{135}^{\mathrm{haze}}\}\).

In our implementation, haze is generated with fixed parameters:
\begin{equation}
\begin{aligned}
A^\infty&=0.90,\quad \beta_{\mathrm{scat}}=2.50,\\
P^A&=0.15,\quad P^T=0.08,\quad \phi_{\mathrm{ref}}=0^\circ,
\end{aligned}
\end{equation}
where \(A^\infty\) is the atmospheric light, \(P^A\) and \(P^T\) denote the degrees of polarization of airlight and transmitted light, and \(\phi_{\mathrm{ref}}\) is the reference polarization direction.

\section{Physical Consistency Visualization} 
\begin{figure}[!t]
  \centering
  \includegraphics[width=\linewidth,trim=4 608.5 243 9,clip]{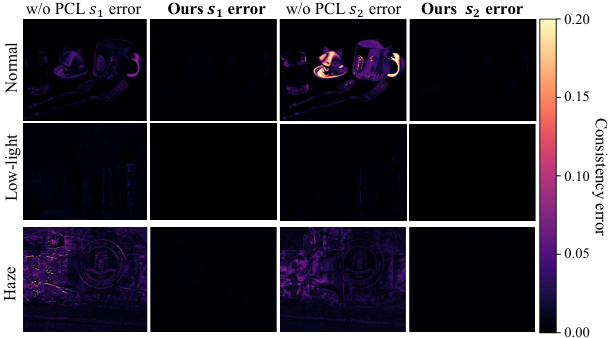}
  \caption{Image--Stokes inconsistency maps under normal, low-light, and haze conditions. Lower values indicate better physical consistency.}
  \label{fig:supp_pcl_vis}
\end{figure}

Figure~\ref{fig:supp_pcl_vis} complements the quantitative analysis in the main paper by visualizing the inconsistency between the image-derived and directly predicted Stokes components. Without PCL, noticeable structured residuals remain in both $s_1$ and $s_2$, particularly around object boundaries and textured regions. After introducing PCL, these residuals are visibly reduced across normal, low-light, and haze conditions, indicating closer agreement between the two output representations. The visualization therefore provides qualitative evidence that PCL improves output-level Image--Stokes physical consistency.

\section{Hyperparameter Analysis}
\label{sec:hyperparameter_analysis}
This section examines the sensitivity of UniPolar to its loss weights, DoLP confidence thresholds, and modulation-scale initialization. These experiments identify settings that balance image fidelity and polarimetric reliability across compound degradations.

\noindent\textbf{Stokes consistency weight $\lambda_1$.}
Table~\ref{tab:lcons_analysis} shows that increasing $\lambda_1$ from 0.1 to 0.2 improves all reported metrics, with $\lambda_1=0.2$ achieving the best mean-image, $s_0$, and DoLP performance. A further increase to 0.3 slightly reduces the AoLP error to 16.796$^\circ$, but degrades the remaining metrics, indicating that an overly strong consistency constraint may compromise restoration fidelity and polarization magnitude estimation. We therefore set $\lambda_1=0.2$.

\begin{table}[!t]
  \caption{Effect of the Stokes consistency loss $\mathcal{L}_{\mathrm{cons}}$ weight $\lambda_1$.}
  \label{tab:lcons_analysis}
  \centering
  \fontsize{8}{8}\selectfont
  \renewcommand{\arraystretch}{1.2}
  \setlength{\tabcolsep}{7pt}
  \setlength{\arrayrulewidth}{0.55pt}
  \arrayrulecolor{tblLine}
  \begin{tabular}{c||cccc}
    \hline
    \textbf{$\lambda_{1}$} & \textit{Mean}$\uparrow$ & $s_0$$\uparrow$ & $p$$\uparrow$ & $\theta$$\downarrow$ \\
    \hline
    \hline
    0.1 & 30.962/0.8655 & 31.874/0.8924 & 23.974/0.5970 & 17.462 \\
    \cellcolor{mygray}0.2
    & \cellcolor{mygray}\textcolor{red}{31.522/0.8752}
    & \cellcolor{mygray}\textcolor{red}{32.446/0.9006}
    & \cellcolor{mygray}\textcolor{red}{24.097/0.6076}
    & \cellcolor{mygray}16.833 \\
    0.3 & 30.975/0.8651 & 32.102/0.8926 & 22.883/0.5841 & \textcolor{red}{16.796} \\
    0.4 & 31.109/0.8663 & 32.071/0.8937 & 23.079/0.5851 & 17.295 \\
    \hline
  \end{tabular}
\end{table}

\begin{table}[!t]
  \caption{Effect of the AoLP loss $\mathcal{L}_{\mathrm{aop}}^w$ weight $\lambda_2$.}
  \label{tab:lambda_aop_analysis}
  \centering
  \fontsize{8}{8}\selectfont
  \renewcommand{\arraystretch}{1.2}
  \setlength{\tabcolsep}{6.5pt}
  \setlength{\arrayrulewidth}{0.55pt}
  \arrayrulecolor{tblLine}
  \begin{tabular}{c||cccc}
    \hline
    \textbf{$\lambda_{2}$} & \textit{Mean}$\uparrow$ & $s_0$$\uparrow$ & $p$$\uparrow$ & $\theta$$\downarrow$ \\
    \hline
    \hline
    0.01 & 30.665/0.8563 & 31.601/0.8851 & 24.075/0.5689 & 17.245 \\
    0.02 & 30.873/0.8615 & 31.799/0.8900 & 23.693/0.5732 & 17.025 \\
    \cellcolor{mygray}0.03
    & \cellcolor{mygray}\textcolor{red}{31.522/0.8752}
    & \cellcolor{mygray}\textcolor{red}{32.446/0.9006}
    & \cellcolor{mygray}\textcolor{red}{24.097/0.6076}
    & \cellcolor{mygray}16.833 \\
    0.04 & 31.175/0.8581 & 32.005/0.8946 & 23.888/0.5942 & 16.657 \\
    0.05 & 30.450/0.8552 & 31.386/0.8833 & 22.868/0.5714 & \textcolor{red}{16.290} \\
    \hline
  \end{tabular}
\end{table}

\begin{table}[!t]
  \caption{Effect of DoLP confidence thresholds $(\tau_l,\tau_h)$.}
  \label{tab:aop_tau_analysis}
  \centering
  \fontsize{8}{8}\selectfont
  \renewcommand{\arraystretch}{1.2}
  \setlength{\tabcolsep}{7pt}
  \setlength{\arrayrulewidth}{0.55pt}
  \arrayrulecolor{tblLine}
  \begin{tabular}{cc||cccc}
    \hline
    \textbf{$\tau_l$} & \textbf{$\tau_h$}
    & All$\downarrow$ & $p>0.05$$\downarrow$ & $p>0.10$$\downarrow$ & $p>0.20$$\downarrow$ \\
    \hline
    \hline
    0.05 & 0.15 & 18.036 & 13.900 & 13.950 & 12.298 \\
    0.03 & 0.20 & 16.936 & 14.438 & 13.320 & 11.605 \\
    \cellcolor{mygray}\textcolor{red}{0.05} & \cellcolor{mygray}\textcolor{red}{0.20}
    & \cellcolor{mygray}\textcolor{red}{16.833}
    & \cellcolor{mygray}\textcolor{red}{13.569}
    & \cellcolor{mygray}\textcolor{red}{12.701}
    & \cellcolor{mygray}\textcolor{red}{11.119} \\
    0.07 & 0.20 & 17.566 & 14.530 & 13.534 & 11.978 \\
    0.05 & 0.25 & 17.928 & 14.889 & 13.880 & 12.195 \\
    \hline
  \end{tabular}
\end{table}

\begin{figure}[!t]
\centering
\includegraphics[width=\linewidth]{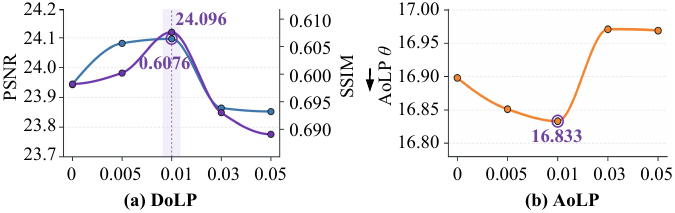}
\caption{Effect of the DoLP feasibility loss $\mathcal{L}_{\mathrm{dop}}$ weight $\lambda_3$.}
\label{fig:hyper-lambda3}
\end{figure}

\noindent\textbf{AoLP loss weight $\lambda_2$.}
Table~\ref{tab:lambda_aop_analysis} shows that increasing $\lambda_2$ from 0.01 to 0.03 improves the overall restoration and polarimetric performance, with $\lambda_2=0.03$ achieving the best mean-image, $s_0$, and DoLP results. Further increasing $\lambda_2$ continues to reduce the AoLP error, reaching 16.290$^\circ$ at $\lambda_2=0.05$, but noticeably degrades image fidelity and DoLP estimation. We therefore set $\lambda_2=0.03$ to balance the objectives.

\noindent\textbf{DoLP confidence thresholds $(\tau_l,\tau_h)$.}
Table~\ref{tab:aop_tau_analysis} shows that $(\tau_l,\tau_h)=(0.05,0.20)$ achieves the lowest angular error over all pixels and across all evaluated DoLP ranges. A smaller $\tau_l$ introduces less reliable weak-polarization directions, whereas larger $\tau_l$ or $\tau_h$ reduces supervision for moderately polarized regions. The selected thresholds balance suppression of uncertain AoLP estimates with retention of informative polarization cues.

\noindent\textbf{DoLP feasibility loss weight $\lambda_3$.}
Figure~\ref{fig:hyper-lambda3} shows that increasing $\lambda_3$ to 0.01 improves both DoLP estimation and AoLP accuracy, achieving 24.097/0.6076 and 16.833$^\circ$, respectively. Larger weights degrade both metrics, indicating that excessive feasibility regularization may over-constrain polarization recovery.

\begin{figure}[!t]
\centering
\includegraphics[width=\linewidth]{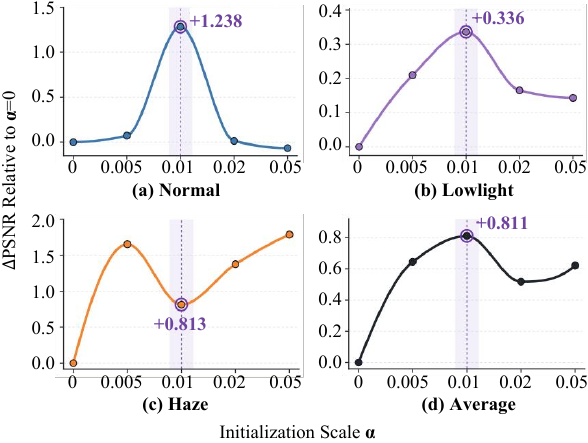}
\caption{Effect of modulation residual-scale initialization $\alpha$ on mean-image PSNR. $\Delta$PSNR is measured relative to $\alpha=0$ under normal, low-light, haze, and average settings.}
\label{fig:hyper-alpha}
\end{figure}

\noindent\textbf{Modulation scale initialization $\alpha$.}
Figure~\ref{fig:hyper-alpha} shows that setting $\alpha=0$ disables the initial contribution of the modulation branch, while a small nonzero value consistently improves performance. In particular, $\alpha=0.01$ yields gains of 1.238, 0.336, and 0.813~dB under normal, low-light, and haze conditions, respectively, and the best average improvement of 0.811~dB. Larger initial scales lead to less stable gains, suggesting that overly strong initial modulation may disturb the backbone features. We therefore initialize $\alpha$ to 0.01.

\bibliographystyle{IEEEtran}
\bibliography{bibliography}

\end{document}